\pdfoutput=1
\documentclass[10pt,twocolumn,letterpaper]{article}
\usepackage[pagenumbers]{cvpr} % To force page numbers, e.g. for an arXiv version
\usepackage{multirow}
\usepackage{booktabs}
\usepackage{amsmath}
\usepackage{colortbl}
\usepackage{tcolorbox}

\usepackage{bm}
\newcommand{\mb}[1]{\mathbf{#1}}
\usepackage{subcaption}
\usepackage{pifont}

\definecolor{cvprblue}{rgb}{0.21,0.49,0.74}
\usepackage[pagebackref,breaklinks,colorlinks,allcolors=cvprblue]{hyperref}

\def\paperID{13070} % *** Enter the Paper ID here
\def\confName{CVPR}
\def\confYear{2025}

\title{Physical Informed Driving World Model}

\author{
Zhuoran Yang $^{1}$\thanks{Equal contributions.}\hspace{0.8em}
Xi Guo$^{2}$\footnotemark[1]\hspace{0.8em}
Chenjing Ding$^{2}$\footnotemark[1] \thanks{Project Leader, Email: \texttt{dingchenjing@sensetime.com}}\hspace{1em}
Chiyu Wang$^2$\hspace{0.8em}
Wei Wu$^{2,3}$\thanks{Corresponding Author, Email: \texttt{wuwei@senseauto.com}}
\\
{
 \normalsize  $^1$University of Science and Technology of China}  \enspace
 \normalsize $^2$SenseAuto \enspace \normalsize $^3$Tsinghua University \\
{{\tt\small shanpoyang@mail.ustc.edu.cn}, \enspace \tt\small \{guoxi,dingchenjing\}@sensetime.com}  \\
{\tt\small \{wangchiyu, wuwei\}@senseauto.com} \\
}

\begin{document}
\maketitle
\begin{abstract}
%The field of autonomous driving increasingly demands high-quality annotated training data. 
% In this paper, we propose \textbf{DrivePhysica}, an innovative approach to generate controllable videos in driving scenarios, capable of yielding an unlimited number of diverse, annotated samples pivotal for autonomous driving advancements.
Autonomous driving requires robust perception models trained on high-quality, large-scale multi-view driving videos for tasks like 3D object detection, segmentation and trajectory prediction. 
While world models provide a cost-effective solution for generating realistic driving videos, challenges remain in ensuring these videos adhere to fundamental physical principles, such as relative and absolute motion, spatial relationship like occlusion and spatial consistency, and temporal consistency.
To address these, we propose \textbf{DrivePhysica}, an innovative model designed to generate realistic multi-view driving videos that accurately adhere to essential physical principles through three key advancements: (1) a Coordinate System Aligner module that integrates relative and absolute motion features to enhance motion interpretation, (2) an Instance Flow Guidance module that ensures precise temporal consistency via efficient 3D flow extraction, and (3) a Box Coordinate Guidance module that improves spatial relationship understanding and accurately resolves occlusion hierarchies.
Grounded in physical principles, we achieve state-of-the-art performance in driving video generation quality ($3.96$ FID and $38.06$ FVD on the Nuscenes dataset) and downstream perception tasks. Our project homepage:
~\href{https://metadrivescape.github.io/papers_project/DrivePhysica/page.html}{https://github.com/DrivePhysica}.
% project web
\end{abstract}    
\section{Introduction}
\label{section:intro}

\begin{figure*}[t]
\centering
\includegraphics[width=0.95\linewidth]{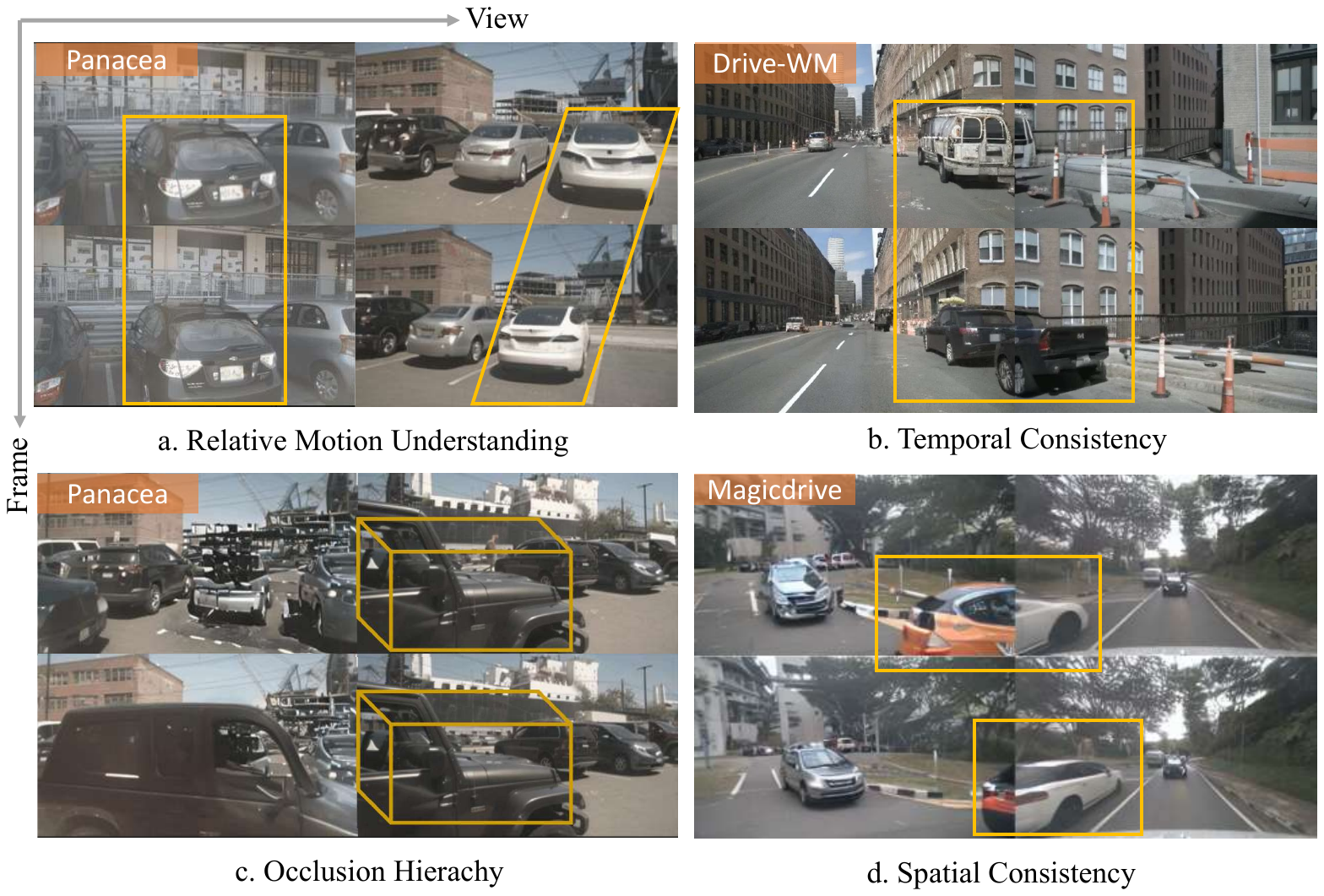} 
\caption{Limitations of previous works in modeling physical laws within driving scenarios. 
\textbf{(a)} Incorrect understanding of relative motion: In reality, the parked black car and the parked white car should exhibit slight movement relative to the ego vehicle. However, the black car remains stationary.
\textbf{(b)} The color of the vehicle changes over time. 
\textbf{(c)} Incorrect understanding of the occlusion hierarchy: The box condition in the background is incorrectly generated in the foreground. 
\textbf{(d)} The appearance of the same car across two views is inconsistent.}
\label{fig:intro}
\end{figure*}

Autonomous driving has attracted extensive attention from both industry and academia for decades \cite{shi2016uniad, zheng2024genad}. To achieve robust perception in autonomous vehicles, models require high-quality, large-scale multi-view driving videos with labeling to train models for tasks like 3D object detection, segmentation, and trajectory prediction. World models \cite{jia2023adriver, wang2023drive} have emerged as a promising solution for generating diverse and realistic driving videos. They can simulate complex scenarios while addressing the high costs and labor of labeling real driving data. 
% which conditions ? list here, describe your task
 % fundamental physical laws

However, generating realistic driving videos that strictly adhere to physical principles—such as relative motion and absolute motion understanding, temporal consistency, and spatial relationship awareness—remains a substantial challenge due to the large sampling space and limited control conditions in diffusion models. Specifically: 
\textbf{1)Motion Reference System Understanding:} Models often struggle to accurately interpret both relative and absolute velocities. For instance, as shown in Fig. \ref{fig:intro} (a), models like Panacea ~\cite{wen2024panacea} fail to understand relative motion. In reality, the parked black car and the parked white car should exhibit slight movement relative to the ego vehicle. However, the black car remains stationary. These limitations in motion comprehension result in unrealistic driving videos, reducing the effectiveness of world models in perception-based tasks. 
\textbf{2)Temporal Consistency:} Preserving stable attributes of moving objects(such as color and texture) over time remains a challenge for numerous driving world models \cite{gao2023magicdrive, ldm, ho2021classifierfree, ho2020denoising, song2020score}. For instance, as depicted in Fig. \ref{fig:intro} (b), DriveWM ~\cite{wang2023drivingfuturemultiviewvisual} fails to maintain temporal consistency, resulting in the car's color varying unrealistically from frame to frame. 
\textbf{3)Spatial Relation Understanding:} Many world models ~\cite{wen2024panacea, gao2023magicdrive} frequently misrepresent spatial relationships, including occlusion hierarchy (correct depth ordering of objects) and cross-view consistency (maintaining coherent structures across multiple camera perspectives). As shown in Fig. \ref{fig:intro} (c), Panacea ~\cite{wen2024panacea} fails to maintain an accurate occlusion hierarchy within vehicle bounding boxes, and the vehicle generated in the bounding box of the rear parking lot was placed closer to the ego vehicle, appearing as a moving vehicle on the road, resulting in the wrong occlusion relationship. 
While in Fig. \ref{fig:intro} (d), MagicDrive ~\cite{gao2023magicdrive} struggles to ensure cross-view spatial coherence. 
%We include additional failure cases from a broader range of methods in the Appendix.

To address these challenges, we propose \textbf{DrivePhysica}, a driving world model that effectively adheres to key physical principles, including motion reference system understanding, temporal consistency, and spatial relationship awareness. DrivePhysica achieves state-of-the-art performance in both the quality of generated videos and validation in downstream tasks.

Firstly, to help model accurately interpret the motion reference system, We introduce \textbf{C}oordinate \textbf{S}ystem \textbf{A}ligner (\textbf{CSA}) module, which uses camera pose parameters to align different conditions under the ego coordinate system and the absolute world coordinate system. 
Camera parameters enable the transformation of absolute world coordinates into the ego-relative coordinate system, thereby aligning the two coordinate systems successfully. 
CSA module provides complementary perspectives, enhancing the model’s understanding of relative and absolute motion.

Secondly, to ensure temporal consistency, we introduce \textbf{I}nstance \textbf{F}low \textbf{G}uidance (\textbf{IFG}) module, a lightweight 3D flow extractor based on the motion vectors of the surrounding instances between frames, avoiding the complex 2D optical flow design used in DrivingDiffusion \cite{li2023drivingdiffusionlayoutguidedmultiviewdriving}.. 
The instance flow serves as a basis to track and propagate attributes (such as color and texture) over frames, aiding in generating temporally consistent videos.
 Operating in the full 3D space instead of being restricted to the 2D image plane, our instance flow enables a more accurate temporal understanding of object positioning.
 
Finally, to help the world model grasp spatial relationships, we propose \textbf{B}ox \textbf{C}oordinate \textbf{G}uidance (\textbf{BCG}) module to embed 3D bounding box coordinates. This direct encoding of 3D positioning helps the model capture occlusion hierarchies. We also ensure cross-view consistency through parameter-free spatial view-inflated attention.

DrivePhysica establishes a robust, physically informed foundation for generating realistic, multi-view driving videos, achieving state-of-the-art performance in both video generation quality and downstream perception task validation. 
Our contributions are three-fold: 
\begin{itemize}
     \item To enhance motion and spatial understanding, we propose a Coordinate System Aligner (CSA) module, enabling the model to better interpret the relationship between absolute and relative motion, addressing a key limitation in current driving world models. By encoding the relative 3D position of each instance in every frame, the model gains enhanced spatial awareness, effectively capturing occlusion hierarchies.
     
    \item To maintain stable object attributes over time, we introduce 3D Instance Flow Guidance with a lightweight flow extractor and operating in 3D space rather than within the 2D image plane, our approach allows for a more precise temporal understanding of object positioning.
    
    %\item We propose the first DiT-based unified control framework DrivePhysica with three novel modules: the Coordinate System Aligner module, the Instance Flow Guidance module, and the Box Coordinate Guidance module. DrivePhysica utilizes these three modules to learn fundamental physical laws in driving scenarios through rich control conditions, enhancing the model’s ability to produce realistic driving simulations.
    
    \item DrivePhysica achieves state-of-the-art (SOTA) performance in both the quality of generated videos and downstream perception metrics. DrivePhysica can simulate long-tail but critical driving scenarios, such as sudden braking and lane cutting, by leveraging synthesized conditions from the Carla Simulator. 
\end{itemize}
\section{Method}
\label{section:method}

In this section, we first define the concept of physical laws that must be satisfied in the context of driving video generation, and then present DrivePhysica, a novel framework for generating realistic multi-view driving videos that faithfully adhere to essential physical principles outlined earlier. 
%The overall architecture is illustrated in Figure \ref{fig: main_figure}.

\begin{figure*}[t]
\centering
\includegraphics[width=1\linewidth]{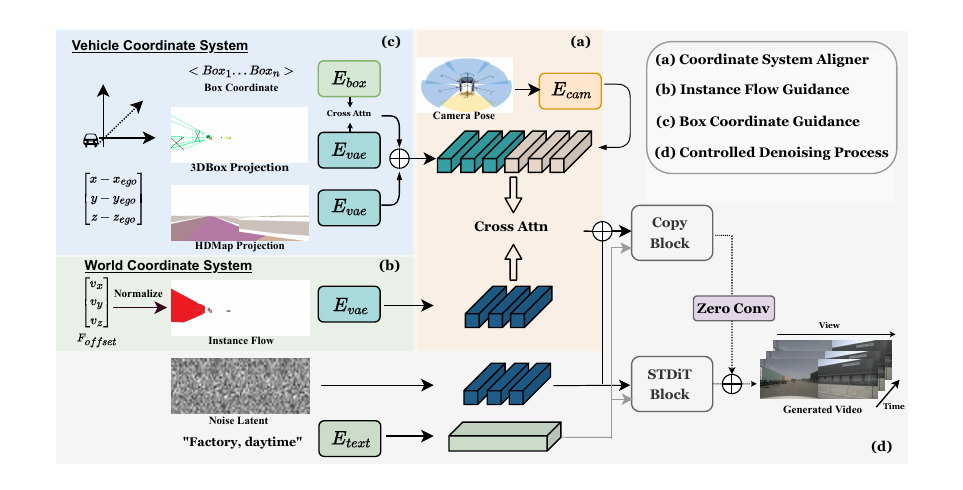} 
\caption{Overview of DrivePhysica.
(a) refers to Coordinate System Aligner module, which uses camera pose parameters to align the vehicle coordinate system with the world coordinate system. 
(b) refers to Instance Flow Guidance module, which utilizes the instance flow to improve temporal consistency.
(c) refers to Box Coordinate Guidance module, which encodes the box coordinates to provide spatial relation information.
(d) refers to Controlled Denoising Process, enabled by ST-DiT with ControlNet for unified condition control.}
\label{fig: main}
\end{figure*}

%-------------------------------------------------------------------------
\subsection{Important Physical Laws to Follow}
\label{section: problem}
In driving video generation, each frame captures a moment. World models for driving scenarios should generate frames that are consistent with real-world physical laws. 
This involves constraints on \textit{motion reference system understanding}, \textit{temporal consistency}, and \textit{spatial relationship awareness}, ensuring that the generated video reflects realistic driving behaviour and physical interactions.

%\begin{itemize}
    \noindent  \textbf{Motion Reference System Understanding.}
    The model must accurately interpret both the world coordinate system and the ego vehicle’s coordinate system. The model must grasp relative motion, where stationary objects in the world coordinate system appear to move in the ego vehicle's coordinate system. Misunderstanding this can lead to erroneous interpretations, such as perceiving a stationary building as moving if the moving ego vehicle is incorrectly assumed to be stationary. 
    As shown in Fig.\ref{fig:intro} (a), the parked cars should exhibit slight movement relative to the ego vehicle, but the black car remains relatively stationary in the ego vehicle’s coordinate system.
    
    \noindent  \textbf{Temporal Consistency.} In accordance with the principles of \textit{conservation of motion} and \textit{material invariance}, object attributes such as color and texture should remain stable across frames. 
    For instance, as depicted in Fig.\ref{fig:intro} (b), the car's color should not vary unrealistically from frame to frame. 

    \noindent  \textbf{Spatial Relationship Awareness.}
    Accurate spatial relationship awareness includes maintaining appropriate distances between objects and properly resolving depth cues to establish a consistent occlusion hierarchy. 
    In a perspective-aligned setup, the relative depth of objects determines their occlusion hierarchy, ensuring that closer objects correctly occlude those further away.
    As shown in Fig.\ref{fig:intro} (c), the condition box in the background is incorrectly generated in the foreground, which obeys occlusion relationship.
    Additionally, the model must avoid introducing discontinuities when transitioning between different camera perspectives, ensuring smooth continuity in the spatial relationships across frames. 
    In Fig.\ref{fig:intro} (d), the car's color should remain consistent across different views.
    %By adhering to these spatial principles, the model ensures that objects are positioned in accordance with physical laws, preserving logical consistency and preventing unrealistic scenarios, such as distant objects appearing in front of closer ones.

%\end{itemize}

\begin{figure*}[t]
\centering
\includegraphics[width=1\linewidth]{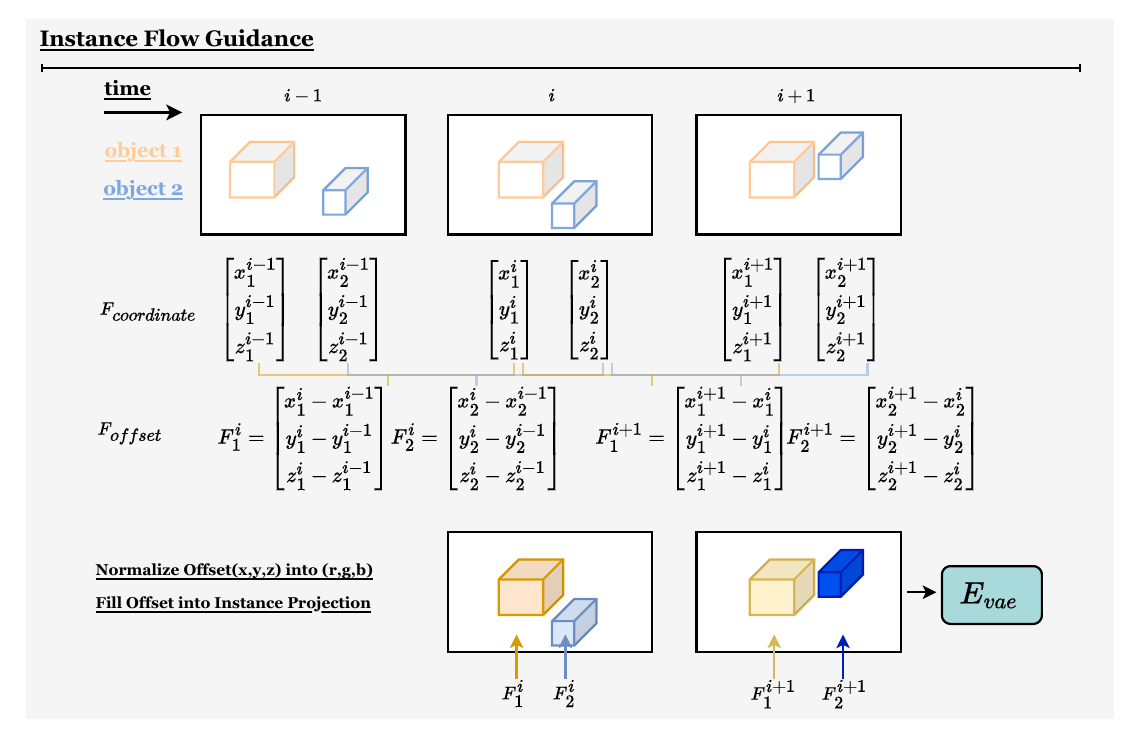} 
\caption{Illustration of the computation process for \textbf{Instance Flow}. 
The process begins by capturing the spatial coordinates of surrounding objects in the absolute world coordinate system over time as \(F_{\text{coordinate}}\). 
Next, the relative motion vectors of these instances between consecutive frames is represented as \(F_{\text{offset}}\). 
Subsequently, the 3D bounding box of each instance is projected onto the camera view to generate its 2D projection area. 
Each pixel in this area at frame \(i\) is populated with the normalized position offset \(F_{\text{offset}}\) of the corresponding instance. 
The offset map is converted into an RGB visualization, encoded into latent space by $E_{vae}$, and integrated into the ST-DiT pipeline, enabling robust and lightweight motion control.}
\label{fig: traj}
\end{figure*}

%-------------------------------------------------------------------------
\subsection{DrivePhysica: Physical Law Acquisition}

To enable the world model to accurately comprehend and adhere to the fundamental physical principles outlined in Sec.\ref{section: problem} and thereby facilitate realistic driving video generation, we integrate three-level control conditions: \textbf{scene condition} (text and camera pose), \textbf{vehicle coordinate system condition} (3D bounding box coordinates, 3D bounding box projection, and road map projection), and \textbf{world coordinate system condition} (instance flow).

Building on OpenSora V1.1 ~\cite{opensora}, we employ a Variational Auto-Encoder (VAE) for video encoding, T5 ~\cite{raffel2020exploring} for text encoding, and Spatial-Temporal Diffusion Transformer (ST-DiT) as the foundational model for the denoising process. 
We reshape the input from $\mathbb{R}^{v \times t \times h \times w \times c}$ to $\mathbb{R}^{t \times h \times (wv) \times c}$ and treat $wv$ as the frame width to improve the consistency of the cross-view.
An overview of the DrivePhysica architecture is presented in Fig. \ref{fig: main}.

%-------------------------------------------------------------------------
\subsubsection{Coordinate System Aligner}
To help the model accurately understand the \textit{motion reference system}, we design the \textbf{C}oordinate \textbf{S}ystem \textbf{A}ligner (\textbf{CSA}) module by aligning features extracted from two essential coordinate systems: \textbf{vehicle coordinate system} and \textbf{world coordinate system}. These two coordinate systems provide complementary perspectives on motion, enabling the model to better understand the relationship between absolute and relative motion.

Previous works typically focus on integrating control conditions based on vehicle coordinate system, such as 3D bounding box projections and road map projections. 
However, only considering conditions based on vehicle coordinate system makes it difficult to accurately comprehend the true motion of objects in world coordinate system, as shown in Fig.\ref{fig:intro}(a).

To address this limitation, we introduce the world coordinate system condition. The necessity of this dual-condition approach arises from the complex dynamics of driving scenes. The vehicle coordinate system condition captures motion relationships relative to the ego-vehicle, while the world coordinate system condition captures forms of motion in the world coordinate system.

However, these two conditions are based on different coordinate systems. 
To accurately generate realistic motion, the model must be capable of comprehending both the world coordinate system and the vehicle coordinate system. 
Understanding the relationship between these two coordinate systems is critical for ensuring that the motion of all scene elements, both foreground and background, adheres to the principles of motion. This alignment significantly enhances the realism of the generated video.

To achieve this, we propose the Coordinate System Aligner module, which uses camera pose parameters, including intrinsic and extrinsic parameters, to align the different conditions in the world coordinate system and vehicle coordinate system. These parameters enable transformation from the world coordinate system into the vehicle coordinate system, thereby aligning the two coordinate systems successfully. Specifically:

\begin{itemize}
    \item \textbf{Vehicle Coordinate System Conditions}, including layout information such as 3D bounding box projections and road map projections, are encoded using Variational Auto-Encoder (VAE) ${E_{vae}}$. 
Bounding box coordinates are encoded using ${E_{box}}$ (we introduce the Box Coordinate Guidance module in Sect. \ref{section: static}). The sum of the three encodings \( h_{\text{map}_{\text{proj}}} \), \( h_{\text{box}_{\text{proj}}} \), and \( h_{\text{coor}} \) is represented as ${h^{vehicle}}$.
These conditions capture the spatial relationships relative to the ego vehicle, align layout information with pixels, and provide essential cues for understanding object positions from the ego vehicle's perspective.
    
    \item \textbf{World Coordinate System Conditions}, such as instance flow, are also encoded through the same VAE ${E_{vae}}$ as ${h^{world}}$ (we introduce Instance Flow Guidance module in Sec. \ref{section: dynamic}). 
%the motion of the surrounding instances between consecutive frames
This condition captures the motion of the surrounding instances between consecutive frames, ensuring continuity across time frames and adherence to physical motion laws, thereby improving temporal consistency.
\end{itemize}

To unify these dual-condition embeddings, the Coordinate System Aligner module leverages the camera pose parameter $\mb{P}=\{\mb{K}\in\mathbb{R}^{3\times 3}, \mb{R}\in\mathbb{R}^{3\times3}, \mb{T}\in\mathbb{R}^{3\times 1}\}$ to facilitate the fusion of the world coordinate system condition and vehicle coordinate system conditions. 
To encode the camera pose parameter, we first concatenate each parameter by columns, resulting in \( \bar{\mb{P}} = [\mb{K}, \mb{R}, \mb{T}]^T \in \mathbb{R}^{7 \times 3} \). 
Since \( \bar{\mb{P}} \) contains values from sine and cosine functions, we apply Fourier embedding ~\cite{mildenhall2020nerf} to each 3-dimensional vector to help the model effectively interpret these high-frequency variations. 
Subsequently, we use a Multi-Layer Perceptron (MLP), denoted \( E_{\text{cam}} \), to embed the camera pose, resulting in the camera embedding \( h^c \):
\[
h^c = E_{\text{cam}}(\operatorname{Fourier}(\bar{\mb{P}})) = E_{\text{cam}}(\operatorname{Fourier}([\mb{K}, \mb{R}, \mb{T}]^T))
\]
which has the same dimension as video patch embedding.
Finally, we leverage the camera pose embedding \( h^c \) to fuse the vehicle coordinate system condition embedding \( h^{vehicle} \) and the world coordinate system condition embedding \( h^{world} \):
\[
h^{condition} = \text{CrossAttn}(\text{cat}(h^{vehicle}, h^c), h^{world}).
\]

The camera pose plays a crucial role in aligning the dual-condition embeddings, merging them into a cohesive control signal. This unified embedding \(h^{condition}\) is then passed into ControlNet \cite{zhang2023adding}, which generates a consolidated control signal to guide the ST-DiT denoising process, enabling the world model to produce realistic and control-precise driving videos.

%-------------------------------------------------------------------------

\subsubsection{Instance Flow Guidance}
\label{section: dynamic}
To ensure \textit{temporal consistency}, we introduce a lightweight \textbf{I}nstance \textbf{F}low \textbf{G}uidance (\textbf{IFG}) module to assist the model in maintaining the stability of object attributes (such as color and texture) over time, illustrated schematically in Fig.~\ref{fig: traj}.

\noindent\textbf{Instance Flow Representation.}
We introduce the \textit{instance flow} condition, which refers to the motion vectors of surrounding instances in the driving scene. 
We first capture the spatial coordinates of surrounding objects in the absolute world coordinate system across time:  
\[
F_{j_{\text{coordinate}}} = \left \{ (x_{j}^{i}, y_{j}^{i}, z_{j}^{i}) \right \}_{i=0}^{T-1},
\]  
where \((x_{j}^{i}, y_{j}^{i}, z_{j}^{i})\) represents the spatial position of instance \(j\) at frame \(i\). 
To model the motion of the surrounding instances between consecutive frames, we define the \textit{instance flow offset}, which encodes the motion vectors between these frames:  
\[
F_{j_{\text{offset}}} = \left\{ (x_{j}^{i} - x_{j}^{i-1}, y_{j}^{i} - y_{j}^{i-1}, z_{j}^{i} - z_{j}^{i-1}) \right\}_{i=1}^{T}.
\]
The complete instance flow offset for all \( N \) instances at the \( i \)-th frame is defined as:  
\[
F_{\text{offset}}^{i} = \left\{ (x_{j}^{i} - x_{j}^{i-1}, y_{j}^{i} - y_{j}^{i-1}, z_{j}^{i} - z_{j}^{i-1}) \right\}_{j=0}^{N},
\]
which serves as a basis to track and propagate attributes (such as color and texture) over frames, aiding in generating temporally consistent videos.

\noindent \textbf{Filling Instance Offsets into Pixel Positions.}
Direct application of frame-to-frame offsets is incompatible with ST-DiT due to its video autoencoder and patchification process. To address this, we transform \( F_{\text{offset}}^{i} \) into a trajectory map \( h^i \in \mathbb{R}^{H \times W \times 3} \) that aligns with the latent space of video patches.  
The instance’s 3D bounding box is projected onto the camera view to obtain its 2D projection area. Each pixel in this area at frame \(i\) is populated with the position offset of the corresponding instance:  
\[
h^i(h, w) = 
\begin{cases}
F_{j_{\text{offset}}}^i, & \text{if instance } j \text{ projects onto } \\
                             &  (h, w) \text{ at frame } i, \\
0, & \text{otherwise.}
\end{cases}
\]  
The position offsets of all instances at frame \(i\) $F_{\text{offset}}^{i}$ are collectively mapped to the trajectory map \( h^i(h, w) \). For the first frame (\(i=0\)), the map is initialized as a zero matrix: \( h^{i=0}(h, w) = 0 \).  

\noindent \textbf{Normalization and RGB Encoding.}  
The trajectory map \( h \) is normalized and converted into RGB space to create a visualized version:  
\[
h^{vis} = Normalize(h),
\]  
where the \(x\)-offset channel corresponds to the red (\(R\)) channel, the \(y\)-offset channel to green (\(G\)), and the \(z\)-offset channel to blue (\(B\)).

\noindent Using the same video encoder ${E_{vae}}$ from the OpenSora framework, the visualized trajectory map \( h^{vis} \) is encoded into a latent representation:  
\[
h^{world} \in \mathbb{R}^{T \times H \times W \times 4}.
\]  
This latent representation integrates seamlessly into the ST-DiT pipeline, enabling robust and lightweight instance flow control.

%-------------------------------------------------------------------------
\subsubsection{Box Coordinate Guidance}
\label{section: static}
To help the world model understand the \textit{spatial relation}, we introduce a \textbf{B}ox \textbf{C}oordinate \textbf{G}uidance (\textbf{BCG}) module, utilizing 3D bounding box coordinates under the vehicle coordinate system as control conditions. This captures the relative distance of instances from the ego-vehicle, providing depth cues that assist the model in understanding occlusion hierarchy.

The driving scene contains a variable number of 3D bounding boxes. We encode each bounding box \(i\) in each frame \(t\) into a hidden vector \( h^{b_i}_t \), with dimensions that match those of the video patches. A 3D bounding box \( (c^i_t, b^i_t) \) consists of two types of information: the class label \( c^i_t \) and the position of the box \( b^i_t \).
For class labels, we follow a method similar to \cite{li2023gligen}, where the pooled embeddings of the class names, denoted \( L_{c^i_t} \), are used as label embeddings. For the box positions \( b^i_t \in \mathbb{R}^{8 \times 3} \), which are represented by the coordinates of the 8 corner points, we apply Fourier embedding to each point and pass it through an MLP to obtain the encoded position, as described in Equation \ref{equ:box_pos}.
We then use another MLP to combine both the class and position embeddings into a single hidden vector, as shown in Equation \ref{equ:box_hidden}. The final hidden states for all the bounding boxes in frame $t$ are represented as \( h_{{coor}_t} = [h^{b_1}_t, \dots, h^{b_{N_{t}}}_t] \), where \( N_{t} \) is the number of bounding boxes in frame $t$.

\begin{align}
    c^{b_i}_t &= \operatorname{AvgPool}(E_{\text{text}}(L_{c^i_t})), \\
    p^{b_i}_t &= \operatorname{MLP}_p(\operatorname{Fourier}(b^i_t)),
    \label{equ:box_pos}, \\
    h^{b_i}_t &= E_{\text{box}}(c^i_t, b^i_t) = \operatorname{MLP}_b(c^{b_i}_t , p^{b_i}_t).
    \label{equ:box_hidden}
\end{align}

We fuse the three vehicle coordinate system conditions \( h_{\text{map}_{\text{proj}}} \), \( h_{\text{box}_{\text{proj}}} \), and \( h_{\text{coor}} \) using cross-attention, where the box embedding serves as inputs to the attention mechanism, and the coordinate embedding serves as the key-value. The fusion process is represented as:
\[
h_{\text{vehicle}} = h_{\text{map}_{\text{proj}}} + \text{CrossAttn}(h_{\text{box}_{\text{proj}}}, h_{\text{coor}}),
\]
where \( h_{\text{vehicle}} \) represents the resulting {vehicle coordinate system conditions} feature after integrating the three input conditions.

This latent representation \( h_{\text{vehicle}} \) integrates seamlessly into the ST-DiT pipeline, enabling concise control on box coordinate and map projection.
Ideally, through training, the model learns to capture instances' spatial relations and their occlusion hierarchy.

\subsection{Various Scenarios Simulator}
We utilized CARLA's autopilot and map system to define a series of obstacle interaction behaviors for different scenarios, such as cut-in and braking. Subsequently, we obtained intermediate 3D bounding boxes, lane markings, drivable areas, and other relevant information. Using the ego vehicle coordinate system definition and the intrinsic and extrinsic parameters of the camera from nuScenes, we projected this information onto various viewpoints, converting it into the control conditions required by the model. This process enabled the generation of specified scenes. The system efficiently leverages CARLA's waypoint mechanism, allowing us to randomly generate various events across different maps and regions. This capability provides a rich set of control conditions.
\section{Experiment}
\label{section: exp}

\begin{figure*}[t]
\centering
\begin{minipage}{0.50\linewidth}
    \centering
    \includegraphics[width=\linewidth]{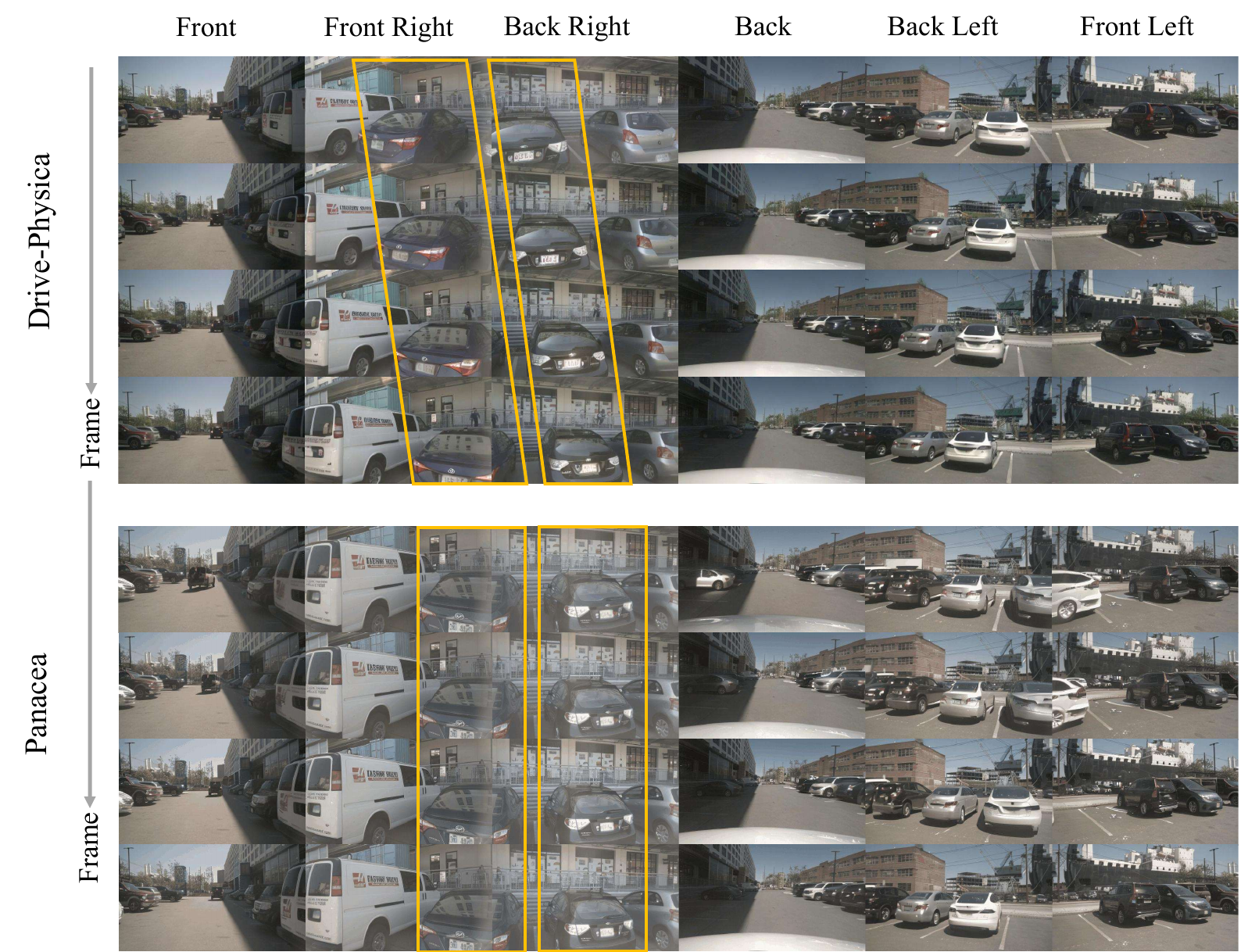} 
    \subcaption{Relative Motion Understanding.}
    %\label{fig:subfig3}
\end{minipage}\hfill
\begin{minipage}{0.50\linewidth}
    \centering
    \includegraphics[width=\linewidth]{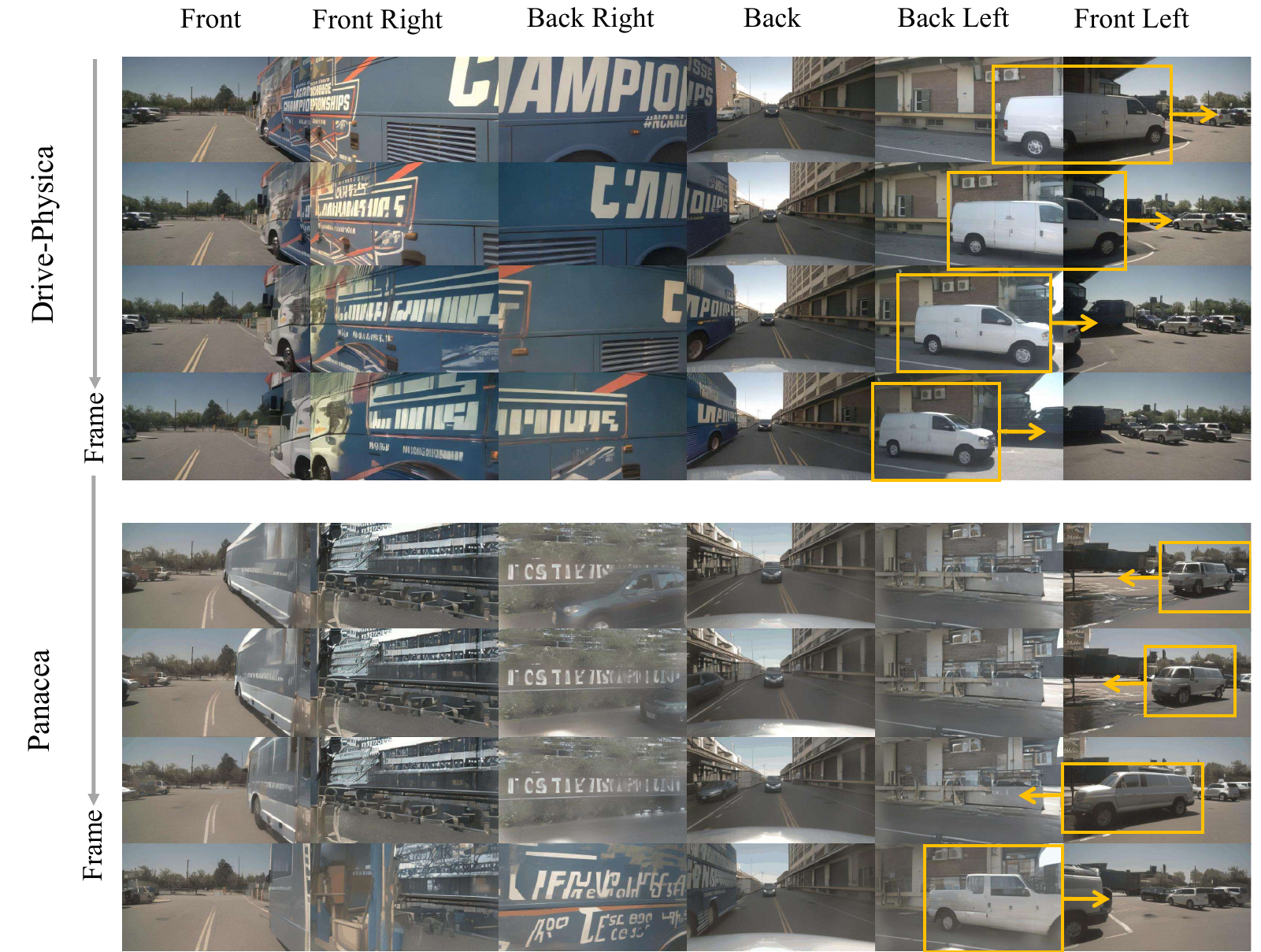} 
    \subcaption{Temporal Consistency.}
    %\label{fig:subfig1}
\end{minipage}\hfill
\begin{minipage}{0.49\linewidth}
    \centering
    \includegraphics[width=\linewidth]{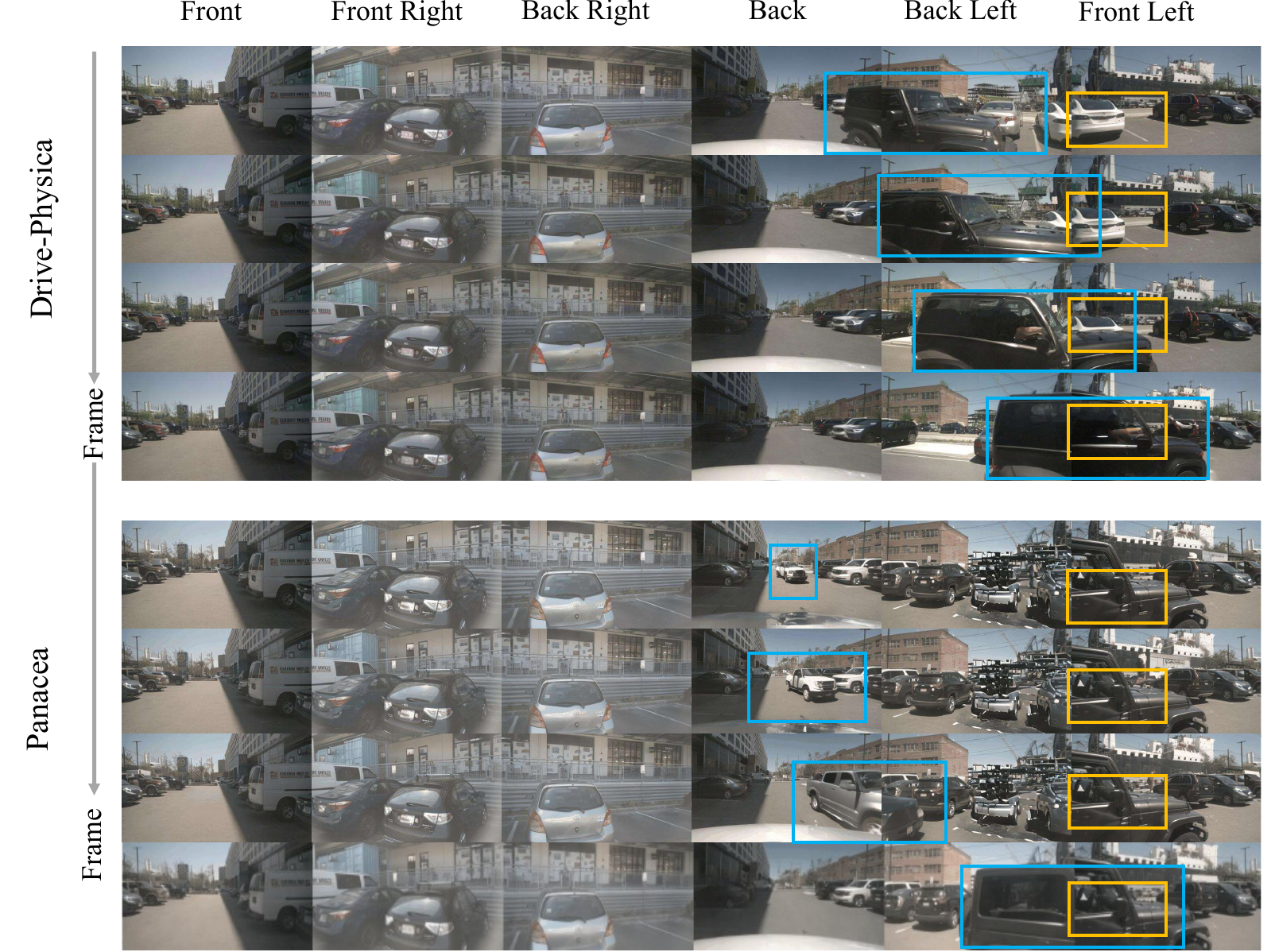} 
    \subcaption{Occlusion Hierarchy.}
    %\label{fig:subfig2}
\end{minipage}
\begin{minipage}{0.49\linewidth}
    \centering
    \includegraphics[width=\linewidth]{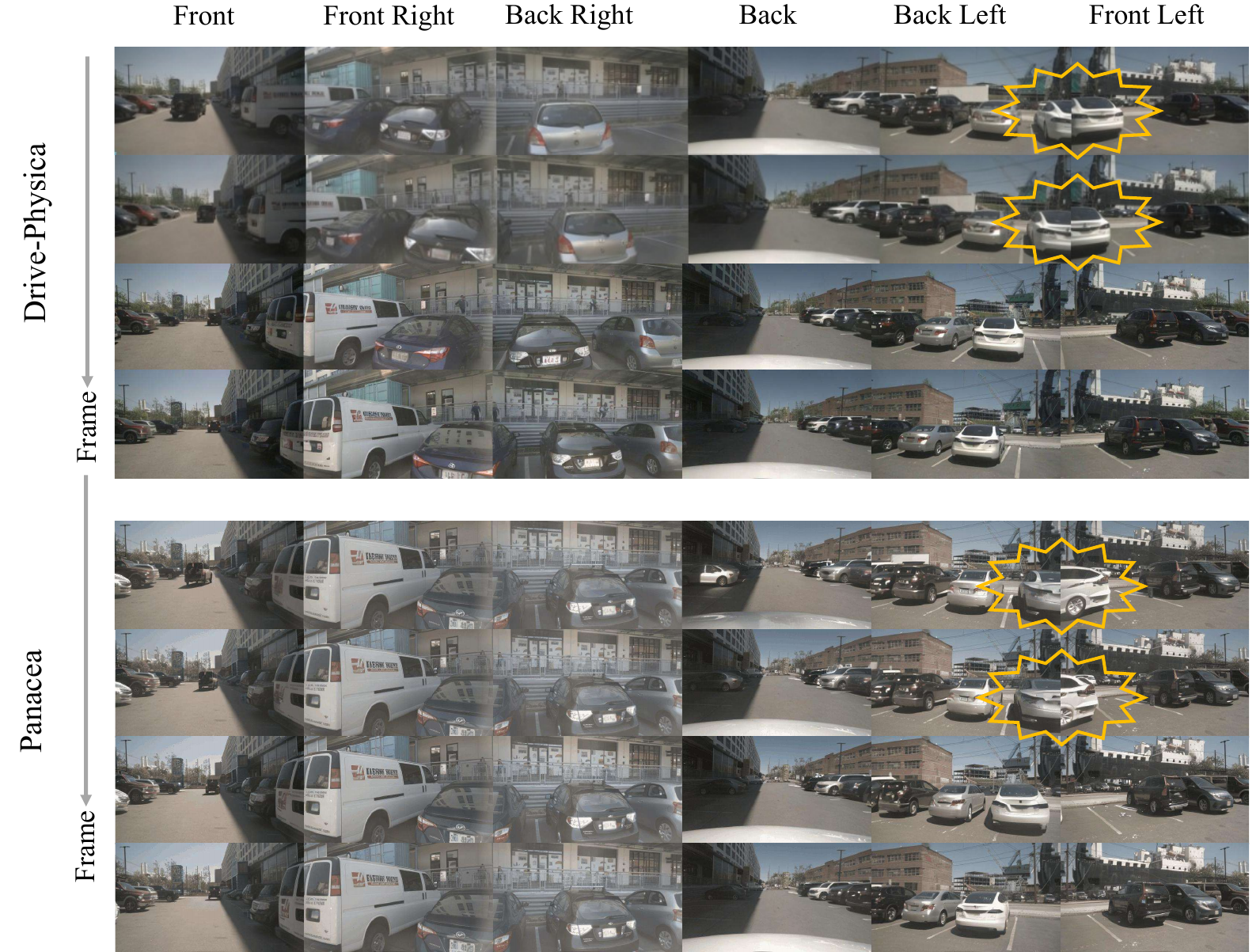} 
    \subcaption{Spatial Consistency.}
    %\label{fig:subfig4}
\end{minipage}
\caption{Qualitative comparison of videos generated by our model and Panacea, the current state-of-the-art method. 
\textbf{(a)} \textbf{Relative Motion Understanding:} As the ego vehicle moves forward, the background and foreground cars should appear to move backward relative to it.  
In Panacea, the black car fails to exhibit correct relative motion and does not move backward as expected relative to the ego vehicle.  
In contrast, our model accurately captures the relative motion of each instance, demonstrating a precise understanding of both the vehicle coordinate system and the world coordinate system.
\textbf{(b)} \textbf{Temporal Consistency:} In Panacea, the white car's shape and orientation (e.g., the direction of the car's front head) change over time. In contrast, our model preserves the white car's attributes throughout the frame, demonstrating superior temporal consistency.
\textbf{(c)} \textbf{Occlusion Hierarchy:} The stationary car (controlled by the orange box condition) is positioned farther from the ego vehicle, while the moving car (controlled by the blue box condition) is closer.  
In Panacea, the generated video incorrectly places the farther stationary car in front, obstructing the closer moving car, therefore violating the expected occlusion hierarchy.  
In contrast, our model correctly renders the closer moving car in front, with the farther stationary car appropriately occluded, demonstrating a superior understanding of occlusion hierarchy.
\textbf{(d)} \textbf{Spatial Consistency:} In Panacea, the white car exhibits different shapes in different views, reflecting spatial inconsistency. 
In contrast, our model maintains consistent spatial representation across views, ensuring coherence throughout the view. 
%Full-length videos are available on our project page in the supplementary materials \textcolor{red}{./drivephysica/page.html}.
}
\label{fig:contrast}
\end{figure*}

\begin{figure*}[t]
\centering
\includegraphics[width=\linewidth]{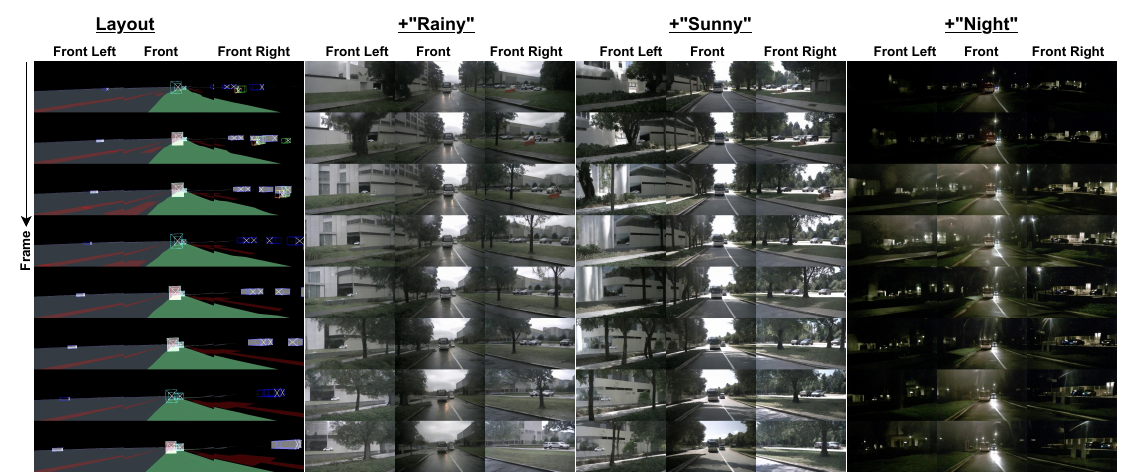} 
\caption{DrivePhysica's \textbf{editing} capability by adjusting the weather and time of day. By adding "Rainy," "Sunny," and "Night" to the original text prompt, while keeping other conditions (such as camera pose, 3D bounding box coordinates, 3D bounding box projections, road map projections, and instance flow) unchanged, our model represents strong ability to edit videos effectively. (a) "Rainy": Captures wet road surfaces and blurred camera views caused by raindrops, adding realistic weather dynamics. (b) "Sunny": Displays clear skies with sunlight shining on the scene, reflecting bright and vivid environmental details. (c) "Night": Depicts dimly lit scenes with streetlights and reduced visibility, accurately simulating nighttime driving conditions. 
%Full-length videos are available on our project page in the supplementary materials \textcolor{red}{./drivephysica/page.html}
}
\label{fig:edit}
\end{figure*}

\begin{figure*}[t]
\centering
\begin{minipage}{\linewidth}
    \centering
    \includegraphics[width=\linewidth]{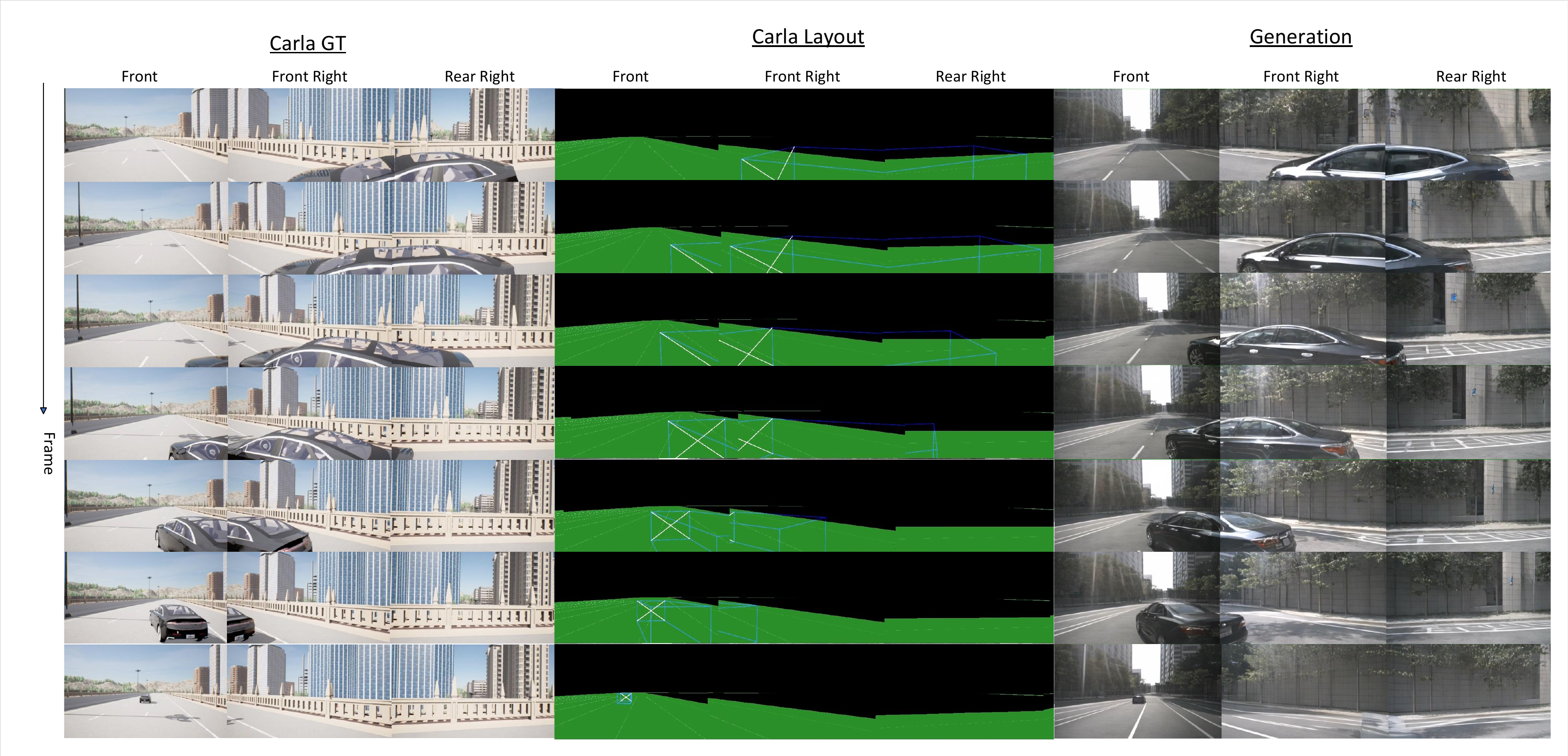} 
    \subcaption{Lane cutting scenario.}
    %\label{fig:subfig4}
\end{minipage}
\begin{minipage}{\linewidth}
    \centering
    \includegraphics[width=\linewidth]{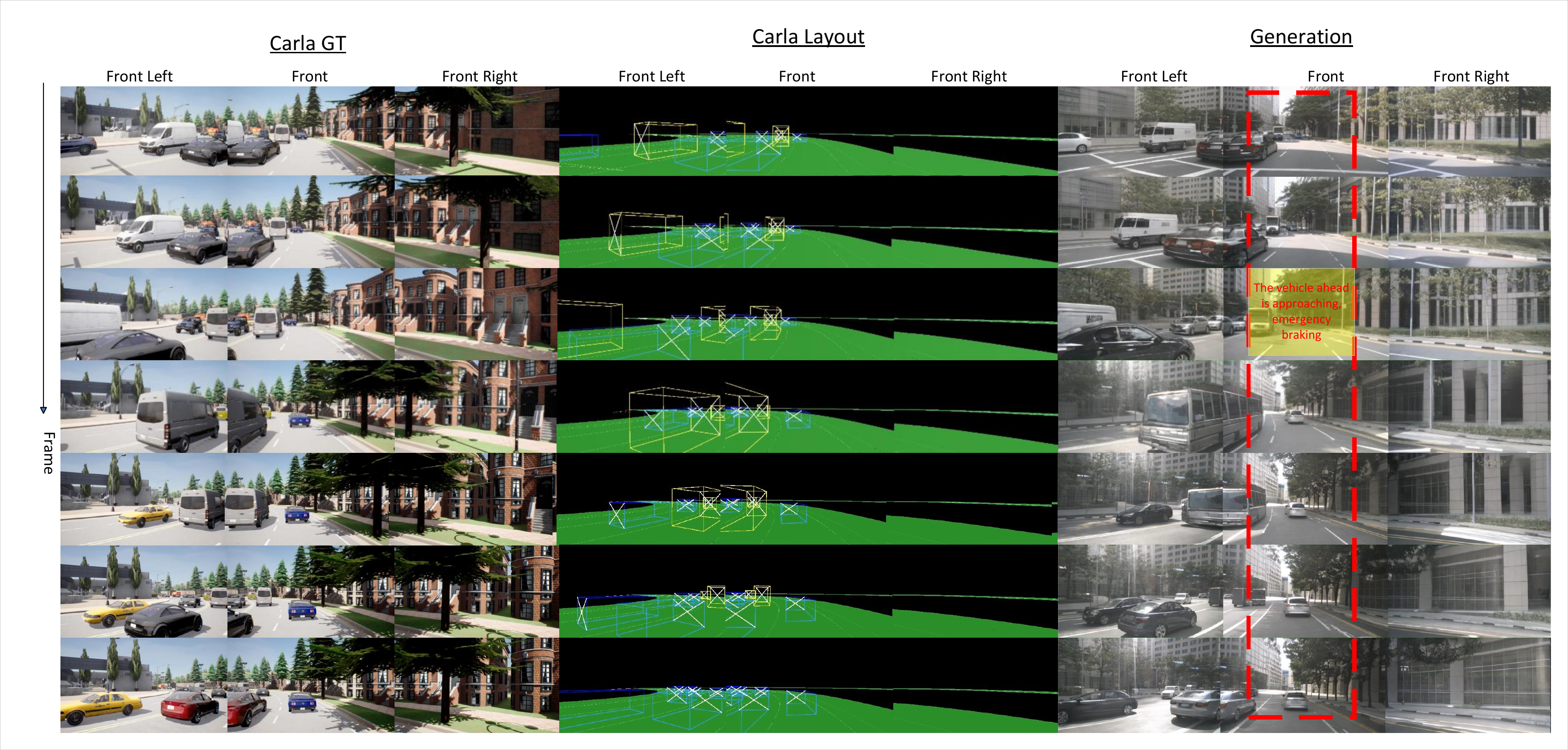} 
    \subcaption{Sudden braking scenario.}
    %\label{fig:subfig5}
\end{minipage}
\caption{DrivePhysica's ability to generate \textbf{rare but critical driving scenarios} based on layout conditions provided by the Carla simulator. The use of Carla-generated layouts addresses a critical limitation in real-world driving video datasets: the lack of diversity in scene types, especially for rare or challenging corner cases. 
% (a) Represents lane cutting scenario. (b) Represents sudden braking scenario.
%Full-length videos are available on our project page in the supplementary materials \textcolor{red}{./drivephysica/page.html}
}
\label{fig:carla}
\end{figure*}

\subsection{Setups}
\noindent \textbf{Datasets and Baselines.} 
We train and evaluate our model using the nuScenes dataset \cite{nuScenes}. We compare our model with image-based solutions (BEVGen \cite{swerdlow2023street}, BEVControl \cite{bevcontrol}) and video-based solutions (DriveDreamer \cite{drivedreamer2}, Panacea \cite{wen2024panacea}).
Our method considers 10 object classes % (car, truck, trailer, bus, construction vehicle, barrier, motorcycle, bicycle, pedestrian, traffic cone) 
and 10 road classes %(drivable area, road segment, road block, lane, ped crossing, walkway, stop line, carpark area, road divider, lane divider)
, surpassing the baseline models in diversity. 

\noindent \textbf{Metrics.}
We evaluate the world model's ability to learn physical laws using FID \cite{FID} and FVD \cite{Thomas2018fvd}. 
The \textit{controllability} of DrivePhysica is demonstrated through the alignment between the generated videos and the conditioned BEV sequences. 
To substantiate this alignment, we evaluate perceptual performance on the nuScenes dataset using metrics such as the nuScenes Detection Score (NDS), mean Average Precision (mAP), mean Average Orientation Error (mAOE), and mean Average Velocity Error (mAVE). We investigate the potential for augmenting the training set to improve model performance. Following Panacea \cite{wen2024panacea}, we use StreamPETR \cite{streampetr}, a state-of-the-art (SoTA) video-based perception method, as the primary evaluation tool.
%Appendix.\ref{app:perception} holds additional details.

\begin{table}[tb]
    \centering
    \footnotesize
        \centering
        \resizebox{0.475\textwidth}{!}{
        \setlength{\tabcolsep}{8pt}
        \begin{tabular}{lcccc}
            \toprule
            Method     &Multi-View &Multi-Frame  & FVD$\downarrow$ & FID$\downarrow$ \\
            \hline
            BEVGen \cite{swerdlow2023street} \  &$\checkmark$   &  &  & 25.54\\
            BEVControl \cite{bevcontrol}  &$\checkmark$   &  & - & 24.85 \\
            WoVoGen \cite{wovogen} &$\checkmark$ & $\checkmark$ & 417.7 & 27.6 \\
            Drive-WM \cite{wang2023drivingfuturemultiviewvisual} &$\checkmark$ & $\checkmark$ & 122.7 & 15.8 \\
            DriveDreamer \cite{wang2023drive}   &$\checkmark$ &$\checkmark$  & 452 & 52.6\\
            Panacea \cite{wen2024panacea}   & $\checkmark$ &$\checkmark$  & 139 & 16.96\\            DriveDreamer2 \cite{drivedreamer2}   &$\checkmark$ &$\checkmark$  & 55.7 & 11.2\\
            \hline
            \rowcolor[gray]{.9} 
            DrivePhysica   &$\checkmark$  &$\checkmark$  & \textcolor{blue}{38.06}  & \textcolor{blue}{3.96}  \\
            \bottomrule
        \end{tabular}
        }
        \caption{Comparing FID and FVD metrics with SoTA methods on the validation set of the nuScenes dataset. We generate the entire validation set without applying any post-processing strategies to select specific samples.}
        \label{tab:fvd}
\end{table}

\subsection{Training Details}
Our method is implemented based on OpenSora \cite{opensora}. Initially, we train for 20k iterations on the front-view videos from the NuScenes training set. Next, to adapt to multi-view positional encoding, we froze the backbone and fine-tuned the patch embedder for 2k iterations. Finally, we added all the control modules and trained the entire model for 100k iterations with a mini-batch size of 1. All training inputs were set to 16x256x448 and conducted on 8 A100 GPUs. Additionally, during training, we set a 0.2 probability of not adding noise to the first frame and assigned a timestep of 0 to the first frame, enabling the model to have image-to-video generation capability. As a result, during testing, the model can autoregressively iterate. Experimental results show that our method can stably generate over 200 frames. For more implementation details 
%and visualization results
, please refer to Appendix.\ref{app: exp}.% and the project webpage included.

% Static 
\subsection{Main Results}
The project webpage ~\href{https://metadrivescape.github.io/papers_project/DrivePhysica/page.html}{https://github.com/DrivePhysica} holds the video demonstration.

\subsubsection{Quantitative Analysis}
\noindent\textbf{Quality of Videos and Adherence to Physical Laws.} To verify the high fidelity of our generated results, we compare our approach with various state-of-the-art driving video generation methods. For fairness, we generate the entire validation set without applying any post-processing strategies to select specific samples. As shown in Tab.~\ref{tab:fvd}, our approach demonstrates significantly superior generation quality, achieving an FVD of $38.06$ and an FID of $3.96$. These metrics substantially outperform those of all counterparts, including both video-based methods like DriveDreamer-2 ~\cite{drivedreamer2} and image-based solutions such as BEVControl ~\cite{bevcontrol}. This highlights that the videos produced by our model exhibit both higher visual quality and better temporal consistency, thereby demonstrating the model's strong capabilities in motion reference system understanding, temporal consistency, and spatial relationship awareness.

\noindent\textbf{Controllability Quality for Autonomous Driving.} 
The controllability of our method is quantitatively assessed through the perception performance metrics derived from the StreamPETR \cite{streampetr} framework. The relative performance metrics, compared to the perception scores of real data, serve as indicators of the control alignment between the generated videos and the control conditions. 

\noindent\textbf{\ding{172}} \textbf{Perception Validation Performance.} We generate the entire validation set of the nuSences by DrivePhysica. 
%The relative performance metrics, compared to the perception scores of real data, serve as indicators of the alignment between the generated samples and the control conditions.
As depicted in Tab.~\ref{tab:valset}, DrivePhysica achieves a relative performance of $86.38\%$ on the nuScenes Detection Score (NDS), employing a pretrained perception model StreamPETR \cite{streampetr}, underscoring a robust alignment between the generated videos and the control conditions. 

\noindent\textbf{\ding{173}} \textbf{Perception Training Performance Using Data Augmentation.} 
DrivePhysica offers the ability to generate an unlimited number of labeled training samples, providing a valuable resource for advancing autonomous driving systems.  
To evaluate this potential, we synthesize a new training dataset specifically tailored for the nuScenes benchmark. This dataset was designed to complement existing training sets for perception models, introducing a stream of diverse and controllable video samples for enhanced learning. 
We retrain the perception model using training datasets with different configurations (real, generated, or a combination of both) and evaluate it on the real nuScenes validation set. The perception metrics are then used to assess how well the generated datasets align with the control conditions.
We first retrain StreamPETR using only real data, setting this as the baseline for comparison, as presented in Tab.~\ref{tab:augment} line 4 (Compared to Panacea in Tab.~\ref{tab:augment} line 1, our re-implemented results on real data are stronger, and under this enhanced model setup, DrivePhysica still demonstrates a significantly greater performance improvement.)
Notably, the perception model trained exclusively on our generated training dataset achieves a mean Average Precision (mAP) of $35.5\%$, equivalent to $92.69\%$ of the performance obtained by models trained solely on the real nuScenes training dataset. 
Similarly, the model achieves a nuScenes Detection Score (NDS) of $43.67\%$, representing $90.41\%$ of the performance achieved using only real data. 
These results highlight that the generated dataset is not only a viable substitute for real data but also highly effective in training perception models independently.
Moreover, by integrating the generated data with real data, the perception model's performance improves significantly, achieving a NDS of $51.9$, marking a $3.6$-point increase over the model trained exclusively on real data. This underscores the substantial value of incorporating generated data into the training pipeline.
Additional comparisons and results for perception testing can be found in Appendix.\ref{app:perception}.

\begin{table}[tb]
    \centering
    \footnotesize
        \centering
        \resizebox{0.475\textwidth}{!}{
        \setlength{\tabcolsep}{7pt}
        \begin{tabular}{cccc}
            \toprule
              Method   & Real & Generated &  NDS$\uparrow$\\
            \hline
                Oracle & $\checkmark$ & -  &46.90 \\ 

                \textbf{Panacea}     & - 
               &$\checkmark$ 
               %&xxx (xx\%) 
               &32.10 (68.00\%) 
               \\
               
               \cellcolor[gray]{.9} \textbf{DrivePhysica}     &\cellcolor[gray]{.9} - 
               &\cellcolor[gray]{.9}  $ \cellcolor[gray]{.9} \checkmark$ 
               %& \cellcolor[gray]{.9} \textcolor{blue}{25.32 (xx\%) }
               & \cellcolor[gray]{.9} \textcolor{blue}{40.51 (86.38\%) }
               \\
            \bottomrule
        \end{tabular}
        }
         \caption{Comparison of the generated data with real data on the nuScenes validation set in (T+I)2V scenarios, employing a pre-trained perception model StreamPETR.
         Our model DrivePhysica achieves a relative performance of $86.38\%$ on the nuScenes Detection Score (NDS), underscoring a robust alignment of the generated samples.}
        \label{tab:valset}
\end{table}

%These results collectively validate that DrivePhysica excels in generating controllable, multi-view driving video samples that can be leveraged to enhance the capabilities of autonomous driving systems. By providing a scalable method to generate high-quality, annotated data, our approach offers significant advantages in training perception models, especially when real data is scarce or difficult to obtain.

% \begin{figure*}[t]
% \centering
% \includegraphics[width=0.9\linewidth]{figures/T2V.pdf} 
% \caption{In T2V scenarios, the generated video showcases the high diversity, controllability, and realism of our method. (To save space, we sample 4 frames from the full video.)}
% \label{fig:t2v}
% \end{figure*}

\subsubsection{Qualitative Analysis}
We present four qualitative comparisons of videos generated by our model and Panacea, evaluating them against the four physical laws defined in Sec. \ref{section: problem}.

\noindent \textbf{Relative Motion Understanding.}
In Fig.\ref{fig:contrast}(a), as the ego vehicle moves forward, the background and foreground cars should appear to move backward relative to it.  
In Panacea, the black car fails to exhibit correct relative motion and does not move backward as expected relative to the ego vehicle.  
In contrast, our model accurately captures the relative motion of each instance, demonstrating a precise understanding of both the vehicle coordinate system and the world coordinate system.

\noindent \textbf{Temporal Consistency.} 
In Fig.\ref{fig:contrast}(b), in Panacea, the white car's shape and orientation (e.g., the direction of the car's front head) change over time. In contrast, our model preserves the white car's attributes throughout the frame, demonstrating superior temporal consistency.

\noindent \textbf{Occulusion Hierarchy.} 
In Fig.\ref{fig:contrast}(c), the stationary car (controlled by the orange box condition) is positioned farther from the ego vehicle, while the moving car (controlled by the blue box condition) is closer.  
In Panacea, the generated video incorrectly places the farther stationary car in front, obstructing the closer moving car, therefore violating the expected occlusion hierarchy.  
In contrast, our model correctly renders the closer moving car in front, with the farther stationary car appropriately occluded, demonstrating a superior understanding of occlusion hierarchy.

\noindent \textbf{Spatial Consistency.} 
In Fig.\ref{fig:contrast}(d), in Panacea, the white car exhibits different shapes in different views, reflecting spatial inconsistency. 
In contrast, our model maintains a consistent spatial representation across views, ensuring coherence throughout the view.

\subsubsection{Long-tail Scenarios Simulation.}
Our DrivePhysica can simulate a variety of long-tail driving scenarios. 
We can change the weather and time of the scenes by modifying the text prompts, as shown in Fig.\ref{fig:edit}. 

We can also generate long-tail but critical events, such as sudden braking and lane cutting, based on the control conditions provided by the \textbf{Carla} simulator \cite{Dosovitskiy17}. It can provide conditions such as 3D bounding box projections, lane line projections, and text descriptions for scenes. 
By leveraging Carla's highly configurable simulation environment, we can create synthetic control conditions that represent complex and diverse driving scenarios, such as multi-vehicle intersections, narrow streets, or sudden obstacles, which are difficult to capture in real-world data.  
In Fig. ~\ref{fig:carla}, we show our model's ability to generate long-tail videos corresponding to these conditions.
The results demonstrate that DrivePhysica can not only replicate realistic conditions, but also seamlessly adapt to a wide range of complex scenarios generated by Carla, further enhancing its applicability in autonomous driving research.

For more details, please refer to Appendix.\ref{app:edit} and Appendix.\ref{app:carla}.

\begin{table}[tb]
    \centering
    \footnotesize
    \resizebox{0.475\textwidth}{!}{
    \setlength{\tabcolsep}{4pt}
    \begin{tabular}{lcccccc}
        \toprule
        Method & Real & Generated & mAP$\uparrow$ & mAOE$\downarrow$ & mAVE$\downarrow$ & NDS$\uparrow$ \\
        \hline
        Panacea & $\checkmark$  & - & 34.5 & 59.4 & 29.1 & 46.9 \\
        % -  &$\checkmark$  &19.7 &74.3 & 49.3 & 34.1 \\
        Panacea & - &$\checkmark$  & 22.5 & 72.7 & 46.9 & 36.1 \\
        %\hline
        %\rowcolor[gray]{.9} 
        % $\checkmark$  & $\checkmark$   & 36.4 \textcolor{blue}{(+1.9\%)} &53.9 &27.8 &48.7 \textcolor{blue}{(+1.8\%)} \\
        Panacea & $\checkmark$  & $\checkmark$   & 37.1 \textcolor{blue}{(+2.6\%)} & 54.2 & 27.3 & 49.2 \textcolor{blue}{(+2.3\%)} \\
        \hline
        DrivePhysica (Re-Implement) &$\checkmark$  & -  & 38.3 & 62.1 & 28.8 & 48.3 \\
        DrivePhysica (Ours) & -  &$\checkmark$  & 35.5 & 59.7 & 29.4 & 43.67 \\
        \rowcolor[gray]{.9} 
        DrivePhysica (Ours) & $\checkmark$  & $\checkmark$   & 42.0 \textcolor{blue}{(+3.7\%)} & 53.2 & 26.8 & 51.9 \textcolor{blue}{(+3.6\%)} \\
        \bottomrule
    \end{tabular}
    }
    \caption{Comparison with Panacea involving data augmentation using synthetic training data to train StreamPETR. We attempt training exclusively using synthetic training data and also explore integrating it with real training data.}
    \label{tab:augment}
\end{table}

\begin{table}[tb]
    \centering
    \footnotesize
        \centering
        \resizebox{0.475\textwidth}{!}{
        \setlength{\tabcolsep}{23pt}
        \begin{tabular}{lcc}
            \toprule
            Settings        & FVD$\downarrow$ & FID$\downarrow$  \\
            \hline
            \rowcolor[gray]{.9} 
            DrivePhysica   & 38.06 & 3.96 \\
            \hline
            w/o Coordinate System Aligner  & 47.34 \textcolor{blue}{(+9.28)}& 5.14 \textcolor{blue}{(+1.18)}\\
            w/o Instance Flow Guidance    & 58.86 \textcolor{blue}{(+20.8)} & 6.28 \textcolor{blue}{(+2.32)}\\ 
            w/o Box Coordinate Guidance & 41.21 \textcolor{blue}{(+3.15)}& 4.22 \textcolor{blue}{(+0.26)}\\           
            \bottomrule
        \end{tabular}
        }
        \caption{Ablation study results in (T+I)2V scenarios on the generated nuScenes validation set.}
        \label{tab: ablation}
\end{table}

\begin{table}[tb]
    \centering
    \footnotesize
        \centering
        \resizebox{0.475\textwidth}{!}{
        \setlength{\tabcolsep}{23pt}
        \begin{tabular}{lcc}
            \toprule
            Settings        & FVD$\downarrow$ & FID$\downarrow$  \\
            \hline
            \rowcolor[gray]{.9} 
            DrivePhysica   & 107.50 & 12.91 \\
            \hline
            w/o Coordinate System Aligner  & 126.08 \textcolor{blue}{(+18.58)}& 15.88 \textcolor{blue}{(+2.97)}\\
            w/o Instance Flow Guidance  & 154.31 \textcolor{blue}{(+46.81)} &  16.73 \textcolor{blue}{(+3.82)}\\ 
            w/o Box Coordinate Guidance & 117.25 \textcolor{blue}{(+9.75)}& 14.52 \textcolor{blue}{(+1.61)}\\           
            \bottomrule
        \end{tabular}
        }
        \caption{Ablation study results in T2V scenarios on the generated nuScenes validation set.}
        \label{tab: ablation2}
\end{table}

\subsection{Ablation Study}
We validate three key modules of DrivePhysica using FID and FVD metrics. Additionally, to assess the effectiveness of these modules, we evaluated the results in two scenarios: with the first frame visible (\textbf{(T+I)2V}) and with the first frame not visible (\textbf{T2V}), as shown in Tabs.\ref{tab: ablation} and \ref{tab: ablation2}.

\noindent \textbf{Coordinate System Aligner.} 
To evaluate the impact of the Coordinate System Aligner module, we conducted an ablation study by removing the camera pose injection. 
In (T+I)2V scenario in Tab.~\ref{tab: ablation}, omitting the camera pose results in a significant degradation of $9.28$ in FVD and $1.18$ in FID.
And in T2V scenario in Tab.~\ref{tab: ablation2}, it results in a degradation of $18.58$ in FVD and $2.97$ in FID.
This underscores the crucial role of the camera pose in aligning different coordinate systems. This alignment is essential for enabling the model to accurately understand and represent the position of instances within the scene. The degradation in both FVD and FID metrics highlights how the absence of this module impairs the model's ability to properly encode spatial and temporal relationships between objects, leading to a loss of motion coherence and visual fidelity in the generated videos.

\noindent \textbf{Instance Flow Guidance.}
To assess the effect of the Instance Flow Guidance module, we conduct an ablation study by eliminating the instance flow injection. 
In (T+I)2V scenario in Tab.~\ref{tab: ablation}, removing the instance flow results in a noticeable drop of $20.8$ in FVD and $2.32$ in FID, highlighting its essential role in maintaining temporal consistency. Moreover, in the T2V mode in Tab.~\ref{tab: ablation2}, where there is no guidance from the first frame, the role of Instance Flow Guidance becomes more significant, and the FVD and FID scores decrease by 46.81 and 3.82 without instance flow injection.
%Furthermore, Fig.~\ref{ablation: flow} clearly demonstrates that without this module, the model struggles to preserve temporal consistency, as evidenced by inconsistencies in the appearance of the car across frames.

\noindent \textbf{Box Coordinate Guidance.}
To evaluate the influence of the Box Coordinate Guidance module, we perform an ablation study by removing the 3D bounding box coordinate condition. 
In (T+I)2V scenario in Tab.~\ref{tab: ablation}, omitting the 3D bounding box condition leads to a significant degradation of $3.15$ in FVD and $0.26$ in FID, emphasizing its vital role in understanding spatial relationships. 
A similar degradation is also observed in the T2V scenario in Tab.~\ref{tab: ablation2}.
%Fig.~\ref{ablation: box} illustrates that, without this condition, the model fails to correctly learn the occlusion hierarchy, as distant vehicles incorrectly occlude closer ones, with the closer car being obscured while the more distant red car remains visible.

To evaluate the impact of three key modules using perception metrics, please refer to Appendix.\ref{app: ablation}.
\section{Conclusion}
\label{section: conclusion}
We propose \textbf{DrivePhysica}, a driving world model that integrates key physical principles to achieve SOTA performance in both video generation and downstream perception tasks. 
Our contributions include the Coordinate System Aligner, 3D Instance Flow Guidance, and Box Coordinate Guidance. 
They enable DrivePhysica to generate high-quality, multi-view driving videos while effectively capturing motion, occlusion hierarchies, and spatial relationships. Our model outperforms existing methods, demonstrating the benefits of incorporating physical principles into driving world models. 

\noindent \textbf{Acknowledegments.} We appreciate the support from Zehuan Wu and Yuxin Guo in the generation of the Carla video and layout sequences.
\clearpage
{
    \small
    \bibliographystyle{ieeenat_fullname}
    \bibliography{main}
}
\clearpage
\appendix
\clearpage
%\setcounter{page}{1}
%\maketitlesupplementary

% \begin{figure*}[t]
% \centering
% \includegraphics[width=1\linewidth]{figures/traj.pdf} 
% \caption{Illustration of the computation process for \textbf{Instance Flow}. 
% The process begins by capturing the spatial coordinates of surrounding objects in the absolute world coordinate system over time as \(F_{\text{coordinate}}\). 
% Next, the relative motion of these instances between consecutive frames is modeled using \(F_{\text{offset}}\), which encodes motion vectors. 
% Subsequently, the 3D bounding box of each instance is projected onto the camera view to generate its 2D projection area. 
% Each pixel in this area at frame \(i\) is populated with the normalized position offset of the corresponding instance. 
% The offset map is converted into an RGB visualization, encoded into latent space, and seamlessly integrated into the ST-DiT pipeline, enabling robust and lightweight motion control.}
% \label{fig: traj}
% \end{figure*}

\section{More Experimental Details}  
\label{app: exp}
%projectpage: \myhref{https://ali-videoai.github.io/tora_video/}

We provide a webpage ~\href{https://github.com/MetaDriveScape/papers_project/tree/main/drivephysica}{https://github.com/DrivePhysica} 
for additional video results.

\section{More Quantitative Results}
\label{app:perception}
We employ StreamPETR ~\cite{streampetr}, a state-of-the-art (SoTA) video-based perception method, as our main evaluation tool.
We compare the validation performance of our generated data against real data.
%The relative performance metrics, compared to the perception scores of real data, serve as indicators of the alignment between the generated samples and the control conditions.

\noindent\textbf{Metrics.} The controllability of \textbf{DrivePhysica} is reflected by the alignment between the generated videos and the conditioned BEV sequences. 
To substantiate this alignment, we assess the perceptual performance on the nuScenes dataset, utilizing metrics such as the nuScenes Detection Score (NDS), mean Average Precision (mAP), mean Average Orientation Error (mAOE), and mean Average Velocity Error (mAVE). 

\subsection{Perception Validation Performance.} 
The controllability of our method is quantitatively assessed based on the perception performance metrics obtained using StreamPETR.  
We generate the entire validation set of the nuSences by \textbf{DrivePhysica}. 
The relative performance metrics, compared to the perception scores of real data, serve as indicators of the alignment between the generated samples and the control conditions.
As depicted in Tab.~\ref{tab:valset}, \textbf{DrivePhysica} achieves a relative performance of $86.38\%$ on the nuScenes Detection Score (NDS), underscoring a robust alignment of the generated samples.

\subsection{Ablation Study on StreamPETR.} 
\label{app: ablation}
We further validate the effectiveness of the three key components of \textbf{DrivePhysica} through perception performance metrics obtained using StreamPETR, evaluated under the (T+I)2V scenario. 
As shown in Tab.~\ref{tab: ablation-appendix}, the perception metrics confirm the contributions of \textbf{Coordinate System Aligner}, \textbf{Instance Flow Guidance}, and \textbf{Box Coordinate Guidance}. 
Among these, \textbf{Instance Flow Guidance} demonstrates the greatest improvement in perception performance, indicating that this module plays the most significant role in enabling the model to generate controllable driving videos. 
This conclusion is consistent with the findings in Tab.\ref{tab: ablation}, which also highlights the critical importance of \textbf{Instance Flow Guidance} in enhancing the model's generative capabilities.

\begin{comment}
\begin{table}[tb]
    \centering
    \footnotesize
        \centering
        \resizebox{0.475\textwidth}{!}{
        \setlength{\tabcolsep}{7pt}
        \begin{tabular}{cccc}
            \toprule
              Stage   & Real & Generated &  NDS$\uparrow$\\
            \hline
                & $\checkmark$ & -  &46.90 \\ 

                \textbf{Panacea}     & - 
               &$\checkmark$ 
               %&xxx (xx\%) 
               &32.10 (68.00\%) 
               \\
               
               \cellcolor[gray]{.9} \textbf{DrivePhysica}     &\cellcolor[gray]{.9} - 
               &\cellcolor[gray]{.9}  $ \cellcolor[gray]{.9} \checkmark$ 
               %& \cellcolor[gray]{.9} \textcolor{blue}{25.32 (xx\%) }
               & \cellcolor[gray]{.9} \textcolor{blue}{40.51 (86.38\%) }
               \\
            \bottomrule
        \end{tabular}
        }
         \caption{Comparison of the generated data with real data on the nuScenes validation set in (T+I)2V scenarios, employing a pre-trained perception model StreamPETR.
         Our model \textbf{DrivePhysica} achieves a relative performance of $86.38\%$ on the nuScenes Detection Score (NDS), underscoring a robust alignment of the generated samples.}
        \label{tab:valset}
\end{table}
\end{comment}

\begin{table}[tb]
    \centering
    \footnotesize
        \centering
        \resizebox{0.475\textwidth}{!}{
        \setlength{\tabcolsep}{7pt}
        \begin{tabular}{ccccc}
            \toprule
              Stage   & Real & Generated  &  NDS$\uparrow$\\
            \hline
                & $\checkmark$ & -   &46.90 \\ 
 
               \cellcolor[gray]{.9} \textbf{DrivePhysica}     &\cellcolor[gray]{.9} - 
               &\cellcolor[gray]{.9}  $ \cellcolor[gray]{.9} \checkmark$ 
               %& \cellcolor[gray]{.9} \textcolor{blue}{25.32 (xx\%) }
               & \cellcolor[gray]{.9} \textcolor{blue}{40.51 (86.38\%) }
               \\
               w/o Coordinate System Aligner     & - 
               &$\checkmark$ 
               %&xxx (xx\%) 
               &38.64 (82.39\%) 
               \\

               w/o Instance Flow Guidance     & - 
               &$\checkmark$ 
               %&xxx (xx\%) 
               &37.31 (79.55\%) 
               \\
               w/o Box Coordinate Guidance      & - 
               &$\checkmark$ 
               %&xxx (xx\%) 
               &40.23 (85.78\%) 
               \\
            \bottomrule
        \end{tabular}
        }
         \caption{Ablation study results in (T+I)2V scenarios, assessed based on the perception performance metrics obtained using StreamPETR.}
        \label{tab: ablation-appendix}
\end{table}

%\input{figures/appendix/comparision-appendix}

% \begin{figure*}[t]
% \centering
% \includegraphics[width=\linewidth]{figures/appendix/edit.pdf} 
% \caption{\textbf{DrivePhysica}'s editing capability by adjusting the weather and time of day. By adding "Sunny," "Rainy," and "Night" to the original text prompt, while keeping other conditions (such as camera pose, 3D bounding box coordinates, 3D bounding box projections, road map projections, and instance flow) unchanged, the high-quality generated videos highlight our model's strong ability to edit videos effectively. (a) "Sunny": Displays clear skies with sunlight shining on the scene, reflecting bright and vivid environmental details. (b) "Rainy": Captures wet road surfaces and blurred camera views caused by raindrops, adding realistic weather dynamics. (c) "Night": Depicts dimly lit scenes with streetlights and reduced visibility, accurately simulating nighttime driving conditions. Full-length videos are available on our project page in the supplementary materials \textcolor{red}{./drivephysica/page.html}.}
% \label{fig:edit}
% \end{figure*}

\begin{figure*}[t]
\centering
% \begin{minipage}{\linewidth}
%     \centering
%     \includegraphics[width=\linewidth]{figures/appendix/control1.pdf} 
%     \subcaption{Relative Motion Understanding.}
%     \label{fig:subfig3}
% \end{minipage}\hfill
\begin{minipage}{\linewidth}
    \centering
    \includegraphics[width=\linewidth]{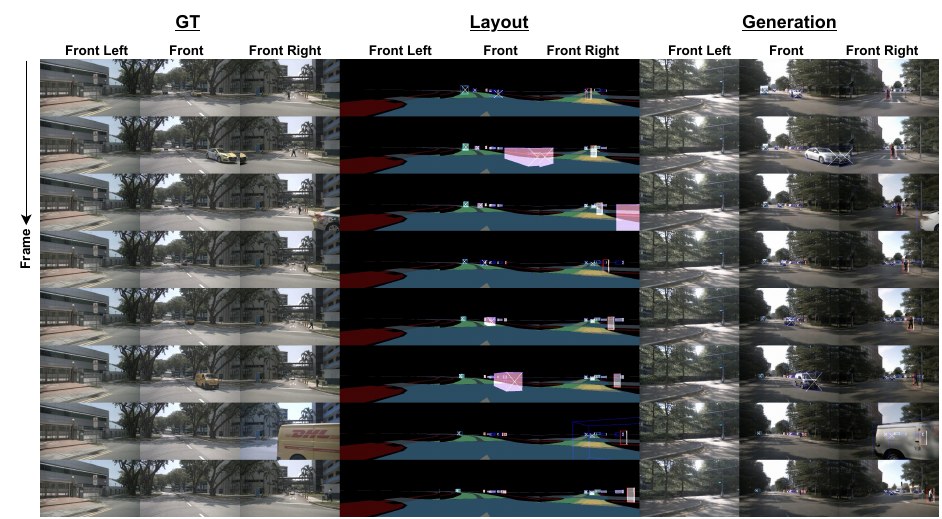} 
    \subcaption{Objects track their attributes maintaining temporal consistency.}
    \label{fig:subfig1_control}
\end{minipage}
\begin{minipage}{\linewidth}
    \centering
    \includegraphics[width=\linewidth]{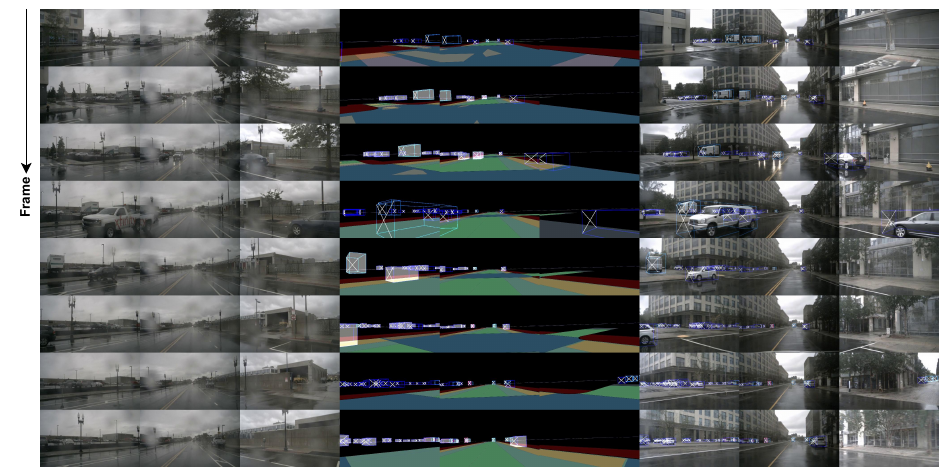} 
    \subcaption{Small and densely packed objects rendered at correct locations.}
    \label{fig:subfig2_control}
\end{minipage}
\caption{ Precise control mechanisms. We overlay the 3D bounding box projections onto the generated videos. The precision of control is reflected in: 
\textbf{(1)} Objects in the scene are accurately placed and sized to align with their \textit{projected bounding boxes}, as shown in \ref{fig:subfig1_control} and \ref{fig:subfig2_control}. 
\textbf{(2)} Drivable areas, sidewalks, and zebra crossings are faithfully generated following the \textit{road map projections}, as shown in \ref{fig:subfig1_control} and \ref{fig:subfig2_control}. 
\textbf{(3)} Objects track their previous attributes as guided by the \textit{instance flow}, ensuring temporal consistency across frames. As shown in Figure~\ref{fig:subfig1_control}, the pink-rendered instance flow directs the model to generate the white sedan, maintaining its consistent attributes over time.
\textbf{(4)} Small and densely packed objects are precisely rendered at their correct locations, following \textit{3D bounding box coordinates}, as shown in \ref{fig:subfig2_control}.
%Full-length videos are available on our project page in the supplementary materials \textcolor{red}{./drivephysica/page.html}
}
\label{fig:control}
\end{figure*}

\begin{figure*}[t]
\centering
\includegraphics[width=\linewidth]{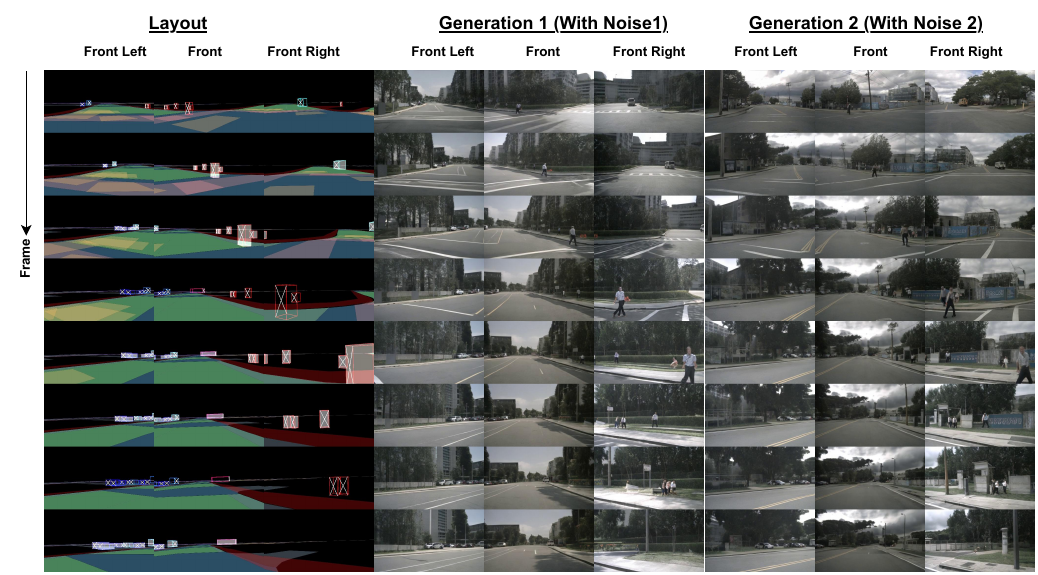} 
\caption{Diverse videos using \textbf{varying noise inputs} and \textbf{the same control conditions}. 
By introducing stochastic noise while maintaining consistent control signals—such as 3D bounding box coordinates, lane line projections, and instance flow—our model can produce a variety of videos that adhere to the defined constraints.
%Full-length videos are available on our project page in the supplementary materials \textcolor{red}{./drivephysica/page.html}
}
\label{fig:noise}
\end{figure*}

\section{More Visualization Results}
Here, we provide additional visualization results to showcase our model's strong ability to generate high-fidelity, realistic, and diverse multi-view driving videos. 
We sample 8 frames from each generated video as a demo to save space in the paper. Our model is capable of generating high-quality, long-duration driving videos through iterative processing. 
We provide a web page in the supplementary materials for additional results. Please refer to the webpage ~\href{https://github.com/MetaDriveScape/papers_project/tree/main/drivephysica}{https://github.com/DrivePhysica} 
.

\subsection{Prompt Edit}
\label{app:edit}
\textbf{DrivePhysica} enables video editing by modifying only the text prompt condition while keeping all other conditions fixed.  
In Fig.~\ref{fig:edit}, we demonstrate the model's editing capability by altering the weather and time of day in the text prompt. Specifically, we add "Sunny," "Rainy," and "Night" to the original text prompt, while maintaining other conditions such as camera pose, 3D bounding box coordinates, 3D bounding box projections, road map projections, and instance flow unchanged. The generated videos showcase high quality and effective editing:  
\begin{itemize}
    \item \textbf{Sunny}: Displays clear skies with sunlight shining on the scene, reflecting bright and vivid environmental details.  
    \item \textbf{Rainy}: Captures wet road surfaces and blurred camera views caused by raindrops, adding realistic weather dynamics.  
    \item \textbf{Night}: Depicts dimly lit scenes with streetlights and reduced visibility, accurately simulating nighttime driving conditions.  
\end{itemize}
These results emphasize the strong editing capability of \textbf{DrivePhysica}, producing diverse and realistic driving videos with minimal changes to the input conditions.

% \textbf{DrivePhysica}'s editing capability by adjusting the weather and time of day. By adding "Sunny," "Rainy," and "Night" to the original text prompt, while keeping other conditions (such as camera pose, 3D bounding box coordinates, 3D bounding box projections, road map projections, and instance flow) unchanged, the high-quality generated videos highlight our model's strong ability to edit videos effectively. (a) "Sunny": Displays clear skies with sunlight shining on the scene, reflecting bright and vivid environmental details. (b) "Rainy": Captures wet road surfaces and blurred camera views caused by raindrops, adding realistic weather dynamics. (c) "Night": Depicts dimly lit scenes with streetlights and reduced visibility, accurately simulating nighttime driving conditions.  

\subsection{Control Precision}

\textbf{DrivePhysica} excels at generating videos that adhere closely to various control conditions, including 3D bounding box coordinates, 3D bounding box projections, road map projections, and instance flow.  
In Fig.~\ref{fig:control}, we overlay the 3D bounding box projections onto the generated videos to illustrate the precision of our control mechanisms.  The precision of control is reflected in:  
\begin{itemize}
    \item \textbf{Object alignment with bounding box projections}: Objects in the scene are accurately placed and sized to align with their \textit{projected bounding boxes}, as shown in  Fig.~\ref{fig:control} (a)(b).  
    \item \textbf{Road and pedestrian area fidelity}: Drivable areas, sidewalks, and zebra crossings are faithfully generated following the \textit{road map projections}, as shown in  Fig.~\ref{fig:control} (a)(b).  
    \item \textbf{Precision in dense and small objects}: Small and densely packed objects are precisely rendered at their correct locations, following \textit{3D bounding box coordinates}, as shown in  Fig.~\ref{fig:control} (b).  
    \item \textbf{Temporal consistency through instance flow}: Objects track their previous attributes as dictated by the \textit{instance flow}, enabling consistent temporal consistency across frames, as shown in  Fig.~\ref{fig:control} (a).  
\end{itemize}
These results highlight the superior control and fidelity of \textbf{DrivePhysica} in generating realistic and controllable driving videos.

% Precision of \textbf{DrivePhysica}'s control mechanisms. We overlay the 3D bounding box projections onto the generated videos. The precision of control is reflected in: (a) Objects in the scene are accurately placed and sized to align with their \textit{projected bounding boxes}. (b) Drivable areas, sidewalks, and zebra crossings are faithfully generated following the \textit{road map projections}. (c) Small and densely packed objects are precisely rendered at their correct locations, following \textit{3D bounding box coordinates}. (d) Objects track their previous attributes as dictated by the \textit{instance flow}, enabling consistent temporal consistency across frames. 

\subsection{Carla-Generated Layout Control}  
\label{app:carla}
\textbf{DrivePhysica} demonstrates the ability to generate high-quality driving videos based on layout conditions provided by the \textbf{Carla} simulator \cite{Dosovitskiy17}, which include 3D bounding box projections, lane line projections, and scene description text prompts. 
The use of Carla-generated layouts addresses a critical limitation in real-world driving video datasets: the lack of diversity in scene types, especially for rare but critical events like lane cutting and sudden braking. 
By leveraging Carla's highly configurable simulation environment, we can create synthetic layouts that represent complex and diverse driving scenarios, such as multi-vehicle intersections, narrow streets, or sudden obstacles, which are difficult to capture in real-world data.  

In Fig.~\ref{fig:carla}, we showcase our model's ability to generate rare videos corresponding to these layouts.   
The generated videos highlight \textbf{DrivePhysica}'s capacity to faithfully adhere to the control signals while producing realistic outputs. 
Moreover, by effectively handling corner cases, our approach bridges the gap in scene diversity, making it a valuable tool for training and validating driving models under challenging scenarios. 
The results demonstrate that \textbf{DrivePhysica} can not only replicate realistic conditions but also adapt seamlessly to a wide range of complex layouts generated by Carla, further enhancing its applicability in autonomous driving research. 
% \textbf{DrivePhysica}'s ability to generate high-quality driving videos based on layout conditions provided by the \textbf{Carla} simulator. The use of Carla-generated layouts addresses a critical limitation in real-world driving video datasets: the lack of diversity in scene types, especially for rare or challenging corner cases. By effectively handling Carla-generated layouts in corner cases, our model can not only replicate realistic conditions but also adapt seamlessly to a wide range of complex layouts, further enhancing its applicability in autonomous driving research. 

\subsection{Diversity with Varying Noise}
\label{app:noise}

\textbf{DrivePhysica} demonstrates the ability to generate diverse driving videos from identical control conditions with \textbf{varying noise inputs}, as illustrated in Fig.~\ref{fig:noise}. By introducing stochastic noise while maintaining consistent control signals—such as 3D bounding box coordinates, lane line projections, and instance flow—our model produces a variety of plausible video outputs that adhere to the defined constraints.

%\subsection{Ablation Results}

\section{Preliminary}
\label{section:preliminary}

\textbf{Latent Video Diffusion Model~(LVDM).}
The LVDM enhances the stable diffusion model ~\cite{ramesh2022hierarchical} by integrating a 3D U-Net, thereby empowering efficient video data processing. This 3D U-Net design augments each spatial convolution with an additional temporal convolution and follows each spatial attention block with a corresponding temporal attention block. It is optimized by employing a noise-prediction objective function:
\begin{equation}
    l_\epsilon = ||\epsilon - \epsilon_\theta(z_t, t, c)||^2_2,
    \label{eq:training_objective}
\end{equation}
Here, $\epsilon_\theta(\cdot)$ signifies the 3D U-Net's noise prediction function. The condition 
$c$ is guided into the U-Net using cross-attention for adjustment. Meanwhile, $z_t$ denotes the noisy hidden state, evolving like a Markov chain that progressively adds Gaussian noise to the initial latent state $z_0$:
\begin{equation}
    z_t=\sqrt{\bar{\alpha}_t}z_0 + \sqrt{1 - \bar{\alpha}_t}\epsilon, \quad\epsilon \sim \mathcal{N}(0, I),
    \label{eq:add_noise}
\end{equation}
where $\bar{\alpha}_t = \prod_{i=1}^t(1-\beta_t)$ and $\beta_t$ is a coefficient that controls the noise strength in step $t$.

\noindent \textbf{Spatial-Temporal DiT (ST-DiT).}
The ST-DiT \cite{peebles2023scalable} introduces a novel architecture that merges the strengths of diffusion models with transformer architectures \cite{vaswani2017attention}. This integration aims to address the limitations of traditional U-Net-based latent diffusion models (LDMs), improving their performance, versatility, and scalability. While keeping the overall framework consistent with existing LDMs, the key shift lies in replacing the U-Net with a transformer architecture for learning the denoising function $\epsilon_\theta(\cdot)$, thereby marking a pivotal advance in the realm of generative modelling.

The ST-DiT architecture incorporates two distinct block types: the Spatial DiT Block (S-DiT-B) and the Temporal DiT Block (T-DiT-B), arranged in an alternating sequence. 
The S-DiT-B comprises two attention layers, each performing Spatial Self-Attention (SSA) and Cross-Attention sequentially, succeeded by a point-wise feed-forward layer that serves to connect adjacent T-DiT-B block. 
Notably, the T-DiT-B modifies this schema solely by substituting SSA with Temporal Self-Attention (TSA), preserving architectural coherence. 
Within each block, the input, upon undergoing normalization, is concatenated back to the block's output via skip-connections. Leveraging the ability to process variable-length sequences, the denoising ST-DiT can handle videos of variable durations.

During processing, a video autoencoder ~\cite{yu2023magvit} is first employed to diminish both spatial and temporal dimensions of videos. To elaborate, it encodes the input video $X \in \mathbb{R}^{T \times H \times W \times 3}$ into video latent $z_{0} \in \mathbb{R}^{t \times h \times w \times 4}$, where $L$ denotes the video length and $t = T, h = H / 8, w = W / 8$.  $z_{0}$ is next ``patchified", resulting in a sequence of input tokens $I \in \mathbb{R}^{t \times s \times d} $. Here, $s = hw/p^2$ and $p$ denote the patch size. 
$I$ is then forwarded to the ST-DiT, which models these compressed representations. 
In both SSA and TSA, standard Attention is performed using Query (Q), Key (K), and Value (V) matrices:
\begin{equation}
Q = W_{Q} \cdot I_{norm}; K = W_{K} \cdot I_{norm}; %V = W_{V} \cdot I_{norm},
\end{equation}
Here, $I_{norm}$ is the normalized $I$, $W_{Q},W_{K}, W_{V}$ are learnable matrices.
The textual prompt is embedded with a T5 encoder and integrated using a cross-attention mechanism.

\section{Related Works}
\label{sec:related}

\noindent \textbf{Controllable Generation.}  
The emergence of diffusion models \cite{zhang2024moonshot} has driven substantial advancements in text-to-video generation ~\cite{an2023latent, blattmann2023align, guo2023animatediff, he2022latent, ho2022imagen, svd, singer2022make, wang2023videofactory, zhou2023magicvideo}. Among these, Video LDM \cite{blattmann2023align} leverages a latent diffusion framework that performs denoising in the image latent space, significantly accelerating the generation process. However, text prompts alone are insufficient for precise video control. Subsequent approaches introduced image blocks in conjunction with textual prompts for the denoising network \cite{zhang2024moonshot}. In our work, we target the generation of highly realistic street-view videos, which present unique challenges due to their complex environments, including intricate street layouts and dynamic vehicles. To achieve fine-grained control, we go beyond text and image inputs by incorporating road maps, 3D bounding boxes, and BEV keyframes, enabling detailed and accurate video generation.

\noindent \textbf{Multi-View Video Generation.}  
Multi-view video generation is challenged by the need for both multi-view consistency and temporal consistency. MVDiffusion \cite{tang2023mvdiffusion} addresses multi-view consistency through a correspondence-aware attention module that aligns information across views. Tseng et al. \cite{poseguideddiffusion} utilize epipolar geometry to enforce view-to-view regularization. Similarly, MagicDrive \cite{gao2023magicdrive} enhances consistency by leveraging priors such as camera poses, bounding boxes, and road maps, along with an additional cross-view attention block. However, these methods primarily focus on generating multi-view images rather than videos and often depend on supplementary data, such as camera poses, which may not be readily available. In contrast, our approach is designed to tackle video generation without reliance on such constraints, offering a more streamlined and efficient solution.

\noindent \textbf{Street-View Generation.}  
Street-view generation methods commonly rely on 2D layouts, including BEV maps, 2D bounding boxes, and semantic segmentation. BEVGen \cite{swerdlow2023street} encodes all semantic information in BEV layouts for street-view generation, while BEVControl \cite{bevcontrol} employs a two-stage pipeline to produce multi-view urban scene images. BEVControl’s controller generates foreground and background objects, while its coordinator ensures cross-view visual consistency. However, the projection of 3D information into 2D layouts results in the loss of geometric details, making these methods prone to temporal inconsistencies when extended to video generation. To address this, we condition generation on 3D bounding boxes, preserving geometric fidelity across frames. While DrivingDiffusion \cite{li2023drivingdiffusionlayoutguidedmultiviewdriving} adopts a multi-stage pipeline involving multiple models and extensive post-processing, our approach simplifies the workflow through an efficient, end-to-end framework, ensuring both temporal coherence and computational efficiency.

\noindent \textbf{Simulation-to-Real Visual Translation.}  
Recent advancements in leveraging synthetic data for real-world visual tasks have seen significant progress across various domains. Notably, methods like GAN-based translation \cite{guo2020gan} and domain randomization \cite{tobin2017domain} have bridged the gap between synthetic and real-world data distributions. Synthetic datasets such as Synthia \cite{ros2016synthia} and Virtual KITTI \cite{cabon2020virtual} have provided scalable benchmarks for semantic segmentation and autonomous driving. Adversarial training approaches \cite{shrivastava2017learning, zhang2018collaborative} have enhanced domain adaptation by reducing distribution discrepancies. Furthermore, human motion representation learning \cite{guo2022learning} has demonstrated the utility of synthetic data in video understanding and biomechanics. These works collectively illustrate the potential of synthetic-to-real transfer in improving model robustness and addressing data scarcity challenges in visual tasks. Unlike these methods, we only extract proxy data such as 3D bounding boxes and road map from the graphics system. Then, utilizing the DrivePhysica model, we can generate more realistic and diverse videos.

\section{Limitations}

Our work establishes a robust, physically informed framework for generating high-quality, multi-view driving videos, achieving state-of-the-art performance in both video generation quality and downstream perception task validation. However, certain limitations remain, mainly due to time and resource constraints.  

Currently, our model's design has not been exhaustively optimized, leaving room for improvement in the quality of the generated videos. For example, the training process is conducted at a relatively low spatial resolution of $256 \times 448$, which constrains visual fidelity. Scaling to higher resolutions would require fine-tuning the position embeddings to ensure compatibility, an aspect not yet addressed in this work.  

Future research could explore the integration of more advanced generative models, such as SD-XL ~\cite{DBLP:journals/corr/abs-2307-01952}, and develop more efficient methods to produce high-fidelity videos at larger spatial resolutions. Additionally, the computational cost of inference for DrivePhysica is relatively high, which presents another avenue for improvement. Enhancing the efficiency of DrivePhysica will be a key focus in future developments to make the model more practical for real-world applications.  

%\clearpage

\end{document}